\documentclass[10pt,twocolumn,letterpaper]{article}
\usepackage{iccv}    
\usepackage{soul}
\usepackage{multirow}
\usepackage{multicol}
\usepackage[table]{xcolor}
\usepackage{amsmath}
\usepackage{bm}
\usepackage{csquotes}
\usepackage{booktabs}
\usepackage{algorithm}
\usepackage{algorithmic}
\usepackage{colortbl}
\usepackage{colortbl}  
\usepackage[accsupp]{axessibility}
\definecolor{customlightgray}{RGB}{230, 230, 230}

\newcommand{\Comment}[1]{\hfill\textit{// #1}}

\definecolor{iccvblue}{rgb}{0.21,0.49,0.74}
\usepackage[pagebackref,breaklinks,colorlinks,allcolors=iccvblue]{hyperref}

\def\paperID{13406} 
\def\confName{ICCV}
\def\confYear{2025}

\title{SpecGuard: Spectral Projection-based Advanced Invisible Watermarking}

\author{
Inzamamul Alam,
Md Tanvir Islam, 
Simon S. Woo\thanks{Corresponding author: swoo@g.skku.edu (Simon S. Woo)}, Khan Muhammad\\
Sungkyunkwan University\\
{\tt\small \{inzi15, tanvirnwu, swoo, khanmuhammad\}@g.skku.edu}
}

\begin{document}
\maketitle
\begin{abstract}
Watermarking embeds imperceptible patterns into images for authenticity verification. However, existing methods often lack robustness against various transformations primarily including distortions, image regeneration, and adversarial perturbation, creating real-world challenges. In this work, we introduce SpecGuard, a novel watermarking approach for robust and invisible image watermarking. Unlike prior approaches, we embed the message inside hidden convolution layers by converting from the spatial domain to the frequency domain using spectral projection of a higher frequency band that is decomposed by wavelet projection. Spectral projection employs Fast Fourier Transform approximation to transform spatial data into the frequency domain efficiently. In the encoding phase, a strength factor enhances resilience against diverse attacks, including adversarial, geometric, and regeneration-based distortions, ensuring the preservation of copyrighted information. Meanwhile, the decoder leverages Parseval’s theorem to effectively learn and extract the watermark pattern, enabling accurate retrieval under challenging transformations. We evaluate the proposed SpecGuard based on the embedded watermark's invisibility, capacity, and robustness. Comprehensive experiments demonstrate the proposed SpecGuard outperforms the state-of-the-art models. To ensure reproducibility, the full code is released on \href{https://github.com/inzamamulDU/SpecGuard_ICCV_2025}{\textcolor{blue}{\textbf{GitHub}}}. 
\vspace{-4mm}
\end{abstract}    

\section{Introduction}
\label{sec:intro}

With the rapid advancement of digital media and artificial intelligence, concerns regarding image authenticity, copyright protection, and content integrity have become more challenging than ever~\cite{pantserev2020malicious, etienne2021future, jayashre2024safeguarding}. Moreover, the widespread availability of the latest image manipulation tools~\cite{goodfellow2014generative, av2024latent, sadre} enables malicious tamperers to easily forge and redistribute digital content without authorization, posing a significant threat to ownership verification~\cite{nguyen2022deep}. This growing risk emphasizes the need for reliable techniques for secure authentication and detection of unauthorized manipulation.

\begin{figure}[!t]
  \centering
   \includegraphics[width=1.0\linewidth]{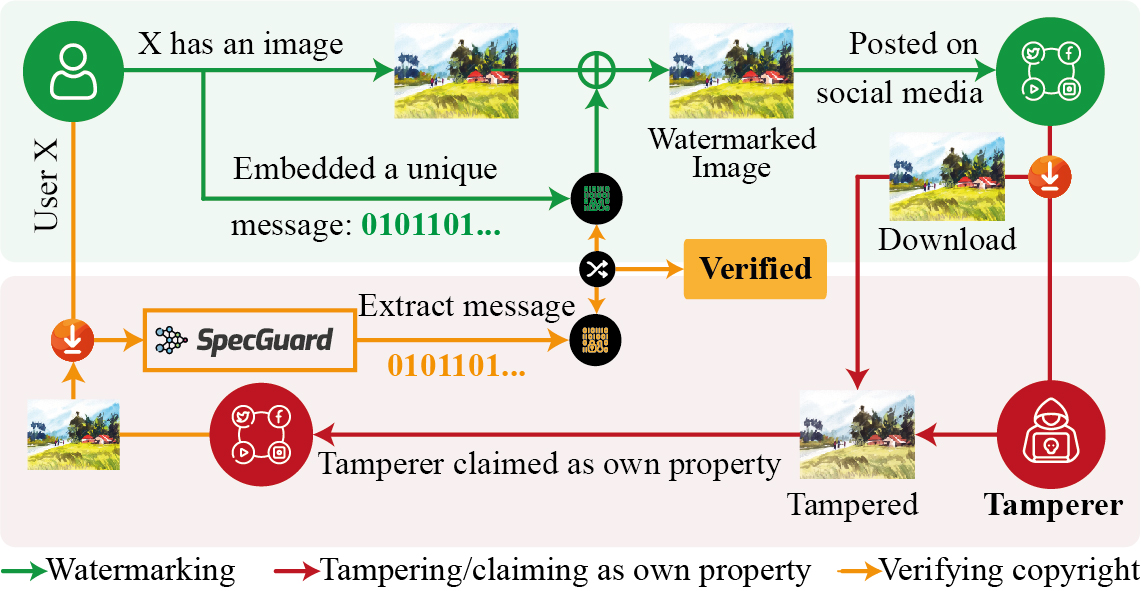}
   \caption{Image authentication using our proposed SpecGuard.}
   \label{fig:motive}
   \vspace{-0.40cm}
\end{figure}

Recently, invisible watermarking has gained significant attention as a prominent defense mechanism for media authentication by embedding invisible messages into images to verify authenticity~\cite{fernandez2023stable, wen2024tree}. In fact, invisible watermarks are preferred for preserving image quality and resisting tampering. These watermarks are unique to the creator and enable tamper verification by comparing the retrieved watermark to the original, as the high-level process is presented in~\cref{fig:motive}. Traditional watermarking methods often rely on transformation techniques~\cite{fft,dwt}. Deep learning approaches like StegaStamp~\cite{tancik2020stegastamp}, Stable Signature~\cite{fernandez2023stable}, and HiDDeN~\cite{hidden2018eccv} provide end-to-end solutions for message embedding. 
\textcolor{black}{However, these methods often struggle with fragility in handling common image processing operations such as resizing, cropping, compression, and noise addition, which can distort or erase the embedded watermark.}
Additionally, the performance of watermark embedding and extraction often remains vulnerable to attacks with noise injection, blurring, contrasting, and rotation~\cite{ding2024waves}.

To address the aforementioned challenges, we introduce a novel robust, and invisible image watermarking method named SpecGuard.  SpecGuard is designed to overcome the fundamental trade-offs~\cite{ding2024waves} between imperceptibility, and robustness. Our proposed SpecGuard strategically embeds watermark information in the spectral domain, leveraging wavelet-based decomposition to distribute the watermark across high-frequency components. Unlike traditional frequency domain watermarking techniques~\cite{yang2024gaussian, NEURIPS2023_b54d1757} that are easily disrupted by common image manipulations, SpecGuard maintains imperceptibility while significantly improving robustness against a wide range of transformations.

Overall, our proposed SpecGuard addresses the current limitations of the previous watermarking methods by providing a robust, imperceptible watermarking technique that maintains integrity under diverse manipulations, significantly enhancing digital content security and authenticity verification. Our key contributions are as follows:

\begin{itemize}
    \item We introduce a novel watermarking approach that embeds message bits in high-frequency spectral components via wavelet and spectral projection inside hidden convolutional layers, ensuring robustness against various transformations and adversarial attacks.

    \item We adapt Parseval’s theorem~\cite{kelkar1983extension} as a learnable threshold to optimize SpecGuard and spectral masking for robust watermark bit recovery under diverse transformations including distortions, regeneration, and adversarial attacks, proven through the experimental results.
    
    \item Our extensive evaluations demonstrate SpecGuard’s superior bit embedding capacity and producing better invisible watermarked images, surpassing the performance of state-of-the-art (SOTA) methods.
\end{itemize}

\section{Related Works}
Watermarking an image has been a widely researched topic for securing the ownership and verifying authenticity of digital content~\cite{sharma2024review}. Traditional watermarking techniques typically embed invisible~\cite{vaishnavi2015robust} or visible~\cite{dekel2017effectiveness} watermarks into images, which can later be extracted or detected to verify the content’s originality. These methods can be broadly classified into spatial-domain~\cite{wang2022color,su2022blind} and frequency-domain~\cite{ghazvini2017improved} watermarking, while some are based on combined methods~\cite{su2020combined,yuan2021blind}. However, researchers recently proposed many advanced models~\cite{liu2021wdnet,niu2023fine,leng2024removing,sun2023denet} for effective watermark removal.
To face this growing challenge, researchers introduced different methods~\cite{hidden2018eccv,luo2020distortion,ahmadi2020redmark,lee2020convolutional,zhang2020udh, SpecXNet, mexfic} as alternatives to deep learning-based encoders or decoders to produce more robust image watermarking. Furthermore, iterative models have demonstrated competitive performance~\cite{kishore2021fixed,fernandez2022watermarking}, particularly in robustness against a wide range of transformations. In addition, with the rise of generative methods, researchers used the watermark-labeled data for training to learn how to produce watermarks~\cite{li2024lightweight,fei2022supervised}. Also, models that combine generative methods with watermarking techniques show promise in effective image watermarking~\cite{lin2024cyclegan,nie2023attributing,qiao2023novel}. However, such approaches face limitations such as increased computational complexity and longer processing times. These approaches are also more vulnerable to adversarial attacks that can target and distort the embedded watermark without altering the content visibly.

\begin{figure*}[ht]
  \centering
   \includegraphics[width=1.0\linewidth]{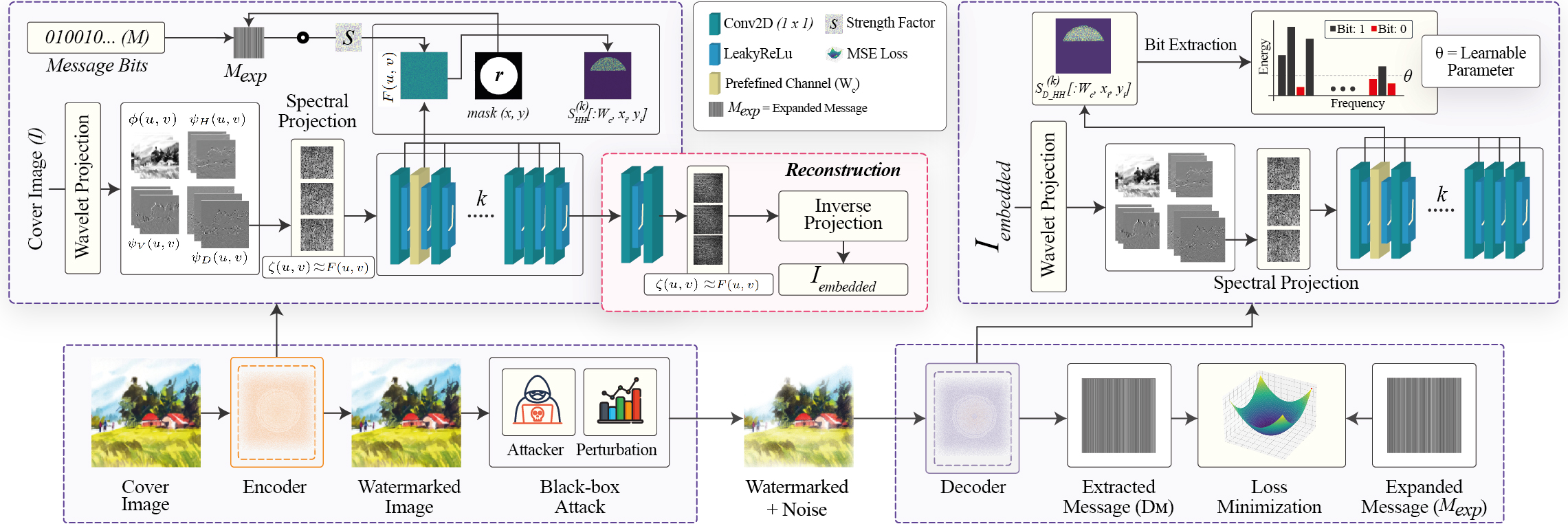}
   \caption{Architecture of the proposed SpecGuard watermarking method involves encoding a binary message $M$ into the high-frequency band of the cover image $I$ using wavelet and spectral projection and learning to decode the embedded message.}
   \label{fig:method}
   \vspace{-0.3cm}
\end{figure*}

\section{Proposed Method: SpecGuard}
We introduce SpecGuard, as illustrated in~\cref{fig:method}, which involves two fundamental modules: an ``Encoder" for embedding the watermark and a ``Decoder'' for accurately extracting the watermark detailed in the following sections.

\subsection{Encoder}
By targeting high-frequency components, the encoder integrates a binary message \( M \) into the cover image \( I \). Using wavelet projection (WP)~\cite{dwt} and a Fast Fourier Transform (FFT)-based spectral projection (SP)~\cite{fft} approximation, the message \( M \) is inserted into specific frequency bands, minimizing perceptual impact.

\noindent
\textbf{Wavelet Projection.}
We use a wavelet projection to capture frequency and spatial localization features that describe an image across different scales, as shown in~\cref{eq:waveletprojection}:
\begin{equation}
\small
 \label{eq:waveletprojection}
    W(a, b) = \frac{1}{\sqrt{|a|}} \int_{-\infty}^{\infty} f(x) \, \psi\left(\frac{x - b}{a}\right) \, dx,
\end{equation}
where \( a \in \mathbb{R} \setminus \{0\} \), and \( b \in \mathbb{R} \) denote the scaling and translation parameters, respectively. Here, \( \psi_{a,b}(x) \) represents a rescaled and translated form of the mother wavelet \( \psi \), defined as follows:
\begin{equation}
\small
    \psi_{a,b}(x) = \psi\left(\frac{x - b}{a}\right) \cdot \frac{1}{\sqrt{|a|}},
\end{equation}
where \( \frac{1}{\sqrt{|a|}} \) functions as a normalization factor, guaranteeing that the energy of the wavelet is invariant to the scaling parameter \( a \). Minimal values of \( a \) compress the wavelet, enabling the inspection of high-frequency components, whereas greater values of \( a \) elongate the wavelet, promoting low-frequency analysis. Since each mother wavelet \( \psi \) is built with zero mean and finite energy~\cite{Fulling1983}, it guarantees to maintain stability as follows:
\begin{equation}
\small
    \int_{-\infty}^{\infty} \psi(x) \, dx = 0, \quad \int_{-\infty}^{\infty} |\psi(x)|^2 \, dx < \infty,
\end{equation}
where the wavelet projection from~\cref{eq:waveletprojection} decomposes the input into orthogonal wavelet sets using discrete scales and translations. For $2D$ inputs, the scaled and translated basis elements~\cite{adelson1987orthogonal} are defined for each coordinate pair \( (u, v) \): 
\begin{equation} \small
\begin{aligned}
\label{eq:frequencybands}
     &\mathbf{S}_{LL} = \mathcal{\phi}(u, v) = \mathcal{\phi}(u)\mathcal{\phi}(v), 
       &\mathbf{S}_{LH} = \mathcal{\psi}_H(u, v) = \mathcal{\psi}(u)\mathcal{\phi}(v), \\
    &\mathbf{S}_{HL} = \mathcal{\psi}_V(u, v) = \mathcal{\phi}(u)\mathcal{\psi}(v), 
      &\mathbf{S}_{HH} = \mathcal{\psi}_D(u, v) = \mathcal{\psi}(u)\mathcal{\psi}(v),
\end{aligned}
\end{equation}
where, \( H \), \( V \), and \( D \) represent the horizontal, vertical, and diagonal decomposition direction, respectively. To depict the image at different resolutions, we define scaling and wavelet functions at scale \( j \) as shown below:
\begin{equation}
\label{eq:imgresolution}
\begin{aligned}
\small
    &\mathcal{\phi}_{j,m,n}(u, v) = 2^{j/2} \mathcal{\phi}\left(u - \frac{m}{2^j}, v - \frac{n}{2^j}\right), \\
    &\mathcal{\psi}_{j,m,n}^d(u, v) = 2^{j/2} \mathcal{\psi}^d\left(u - \frac{m}{2^j}, v - \frac{n}{2^j}\right),
\end{aligned}
\end{equation}
\noindent where \( d \in \{H, V, D\} \) is the wavelet function direction that serves as discrete basis elements for multi-resolution analysis, capturing details across frequency bands and spatial locations. In~\cref{eq:approximation}, \( T_{m,n} \) denotes the intensity or pixel value of the cover image \( I \) at spatial coordinates \( (m, n) \). The discrete scaling function \( W_\phi(j, u, v) \) (approximation at scale \( j \)) and the detail coefficients \( W_{\psi}^d(j, u, v) \) for each direction are computed accordingly as follows: 
\begin{equation} \small
\begin{aligned}
\label{eq:approximation}
    &W_\mathcal{\phi}(j, u, v) = \frac{1}{l} \sum_{m=0}^{l-1} \sum_{n=0}^{l-1} T_{m,n} \, \mathcal{\phi}\left(m - u \cdot 2^{-j}, n - v \cdot 2^{-j}\right), \\
    &W_\mathcal{\psi}^d(j, u, v) = \frac{1}{l} \sum_{m=0}^{l-1} \sum_{n=0}^{l-1} T_{m,n} \, \mathcal{\psi}^d\left(m - u \cdot 2^{-j}, n - v \cdot 2^{-j}\right),
\end{aligned}
\end{equation}
with \( l \) as the discrete region dimension, these coefficients capture multi-scale, multi-orientation image information, forming the basis of spectral features as follows:
\begin{equation}  \small
    \beta_j = \bigcup_{d \in \{H, V, D\}} \left( W_\mathcal{\phi}(j, u, v) \cup W_\mathcal{\psi}^d(j, u, v) \right).
\end{equation}

This feature set $\beta_j$ captures key frequency and spatial details across resolutions, forming the foundation for the watermark embedding process of our SpecGuard.

\noindent
\textbf{Selective Frequency Band Decomposition.}
To refine the embedding process, we segment the data into distinct frequency bands. The decomposition level \( \kappa \) is determined by the image complexity, calculated as follows:
\begin{equation}
\small
    \kappa = \lfloor \sqrt{\log(1 + N)} \rfloor,
\end{equation}
where \( N \) denotes the total pixel count in the cover image \( I \). And, each component \( \beta_j \) falls within a unique frequency band, yielding a total of \( 1 + 3\kappa \) distinct frequency bands as follows:
\begin{equation}
\small
    \beta_j = \phi_j(u, v) \cup \bigcup_{d \in \{H, V, D\}} \psi_j^d(u, v).
\end{equation}

The components \( \beta_j \), consisting of scaling functions \( \phi_j(u, v) \) and wavelet functions \( \psi_j^d(u, v) \), capture specific spatial frequency bands, enabling targeted high-frequency embedding. We translate the WP into disjoint intervals representing a unique frequency range to approximate the segmentation in the frequency domain:
\begin{equation} 
\label{wavelet_componnents}
\small
    \beta_j = \left\{ W_\mathcal{\psi}^d(u, v) \mid u, v \in \left( \frac{j \cdot L}{\kappa}, \frac{(j + 1) \cdot L}{\kappa} \right) \right\},
\end{equation}
where, \( L \) is the dimension of \( S_{HH} \), and \( W_\mathcal{\psi}^d(u, v) \) represents wavelet values within segmented intervals. This frequency band partitioning mimics the frequency selectivity of wavelet sub-bands, enabling effective targeting of high-frequency regions for optimal embedding.

\noindent
\textbf{Approximation of Spectral Projection.}
\textcolor{black}{We first apply spectral projection on the \( S_{HH} \) sub-band, transforming it into the spectral domain. Given a matrix \( T(x, y) \) representing pixel intensities in \( S_{HH} \), the spectral projection computes the spectral components \( \zeta(u, v) \) as follows:
\begin{equation} 
\small
\label{eq:spectral_components}
    \zeta(u, v) = \frac{1}{L^2} \sum_{x} \sum_{y} T(x, y) \cdot \exp\left(-i \frac{2 \pi}{L} \left(x \cdot u + y \cdot v\right)\right),
\end{equation}
where \( L \) denotes the dimension of \( S_{HH} \), \( T(x, y) \) provides the intensity at each coordinate \( (x, y) \) which is equivalent to \( W_\mathcal{\psi}^d(u, v) \) in~\cref{eq:approximation}, \( i \) is the imaginary unit, and \( (u, v) \) are the spectral coordinates.} 

\textcolor{black}{To approximate the spectral components using the FFT, we create a symmetrically extended version \( \tilde{T}(x, y) \) of the original \( N \times N \) matrix \( T(x, y) \). This extension is achieved by mirroring \( T(x, y) \) along its boundaries, doubling its size to \( 2N \times 2N \). Specifically, the original matrix occupies the top-left quadrant, with the remaining quadrants filled by reflecting \( T(x, y) \) horizontally, vertically, and diagonally, respectively. This symmetric structure ensures that the FFT yields only real values, allowing the spectral coefficients to be extracted directly from the real part of the FFT operation. Then, we apply the 2D FFT to \( \tilde{T}(x, y) \) as follows}:
\begin{equation} 
 \small
\label{eq:fft_approximation}
    F(u, v) = \frac{1}{(2N)^2} \sum_{x} \sum_{y} \tilde{T}(x, y) \cdot \exp\left(-i \frac{2 \pi}{2N} \left(x \cdot u + y \cdot v\right)\right).
\end{equation}

The SP coefficients are then approximated by taking the real part (Re) of \( F \) in the original \( N \times N \) region as follows:
\begin{equation}\label{eq:dct_cofficient}
\small
    \zeta(u, v) \approx \operatorname{Re}(F(u, v)), \quad 0 \leq u, v < N.
\end{equation}

\textcolor{black}{Applying~\cref{eq:dct_cofficient} to the sub-bands extracted from wavelet projection in~\cref{eq:approximation}, we achieve a computationally efficient spectral projection by leveraging the FFT approximation on a symmetrically extended matrix, maintaining effective embedding properties within the spectral domain.}

\noindent
\textbf{SpecGuard Embedding Process.}
\textcolor{black}{The embedding process integrates the binary message \( M \) into the high-frequency band \( S_{HH} \) of the cover image \( I \), enhancing robustness and imperceptibility through wavelet and spectral projection. Using the~\cref{eq:approximation} and~\cref{eq:dct_cofficient}, the cover image \( I \) is decomposed into sub-bands \( S_{LL}, S_{LH}, S_{HL}, \) and \( S_{HH} \) within spectral domain, with \( S_{HH} \) providing high-frequency details for embedding. A variable number \( k \) of convolutional layers with a \( K \times K \) kernel, followed by LeakyReLU activation, are recursively applied to \( S_{HH} \) to refine features:}
\begin{equation}  
\label{eq:conv_k}
\footnotesize
S_{HH}^{(n+1)} = \text{LeakyReLU}(\text{Conv}_{2D}(S_{HH}^{(n)}, K)), \quad n = 1, \dots, k.
\end{equation}

The final output \( S_{HH}^{(n+1)} \) from~\cref{eq:conv_k} represents the modified high-frequency band, primed for embedding.




The message \( M \), represented as a binary vector of length \( l \) (\( M \in \{0, 1\}^{l} \)), with batch size \( b \) and message length \( l \), is reshaped and expanded across channels \( c \) to align with \( S_{HH}^{(n+1)} \). This ensures \( M_{\text{expanded}} \) conforms to the dimension \([b, c, l]\), where each message is structured accordingly.

To localize the embedding, we create a radial mask centered at \( (c_x, c_y) = \left( \frac{h}{2}, \frac{w}{2} \right) \), where \( h \) and \( w \) represent the height and width of \( S_{HH} \). The Euclidean distance \( D(x_i, y_i) \)  from the center \((c_x, c_y)\) is computed for each coefficient \( (x_i, y_i) \). A binary mask is then generated within the predefined radius \(r\) based on the distance \( D(x_i, y_i) \), such that if \( D(x_i, y_i) \leq r \), the mask value is set to 1, allowing embedding in the corresponding region. Otherwise, the mask value is 0, restricting embedding to areas within a specified radius \( r \), ensuring focus on high-frequency regions.



For each coordinate \( (x_i, y_i) \) where \( \text{mask }(x_i, y_i)\) is 1 and \( W_{c} \in c \), the embedding operation is performed as follows:
\begin{equation}
\small
S_{HH}^{(n+1)}[:, W_{c}, x_i, y_i] \mathrel{+}= M_{\text{expanded}}[:, W_{c}, i] \cdot s,
\label{strength}
\end{equation}
\noindent where \( s \) is the strength factor controlling embedding intensity and invisibility. After embedding, the modified coefficients \( S_{HH}^{(n+1)} \) undergo a final convolution and LeakyReLU using~\cref{eq:conv_k}, by setting the value of \(k = 1\) to harmonize the embedded message. Following this approach, SpecGuard embeds the message into the spectral domain in a transformed form, differing from its original input representation. By blending the message seamlessly into the spectral space based on the \(r\), \(s\), and \(W_c\), it becomes inherently concealed within the domain, rendering its presence imperceptible. Without knowledge of \(r\), \(s\), and \(W_c\), it becomes exceedingly challenging to localize the embedded message, further enhancing the security of the system. This transformation ensures the embedding process remains opaque to any adversarial attacker,  making SpecGuard black-box.

\noindent
\textbf{Reconstruction.} SpecGuard encoder reconstructs the watermarked image \( I_{\text{embedded}} \) by inverse transformation restoring \( S_{HH} \) back into the spatial domain. The reconstruction process integrates the inverse wavelet projection (IWP)~\cite{dwt} and inverse spectral projection (ISP)~\cite{fft}, ensuring the embedded modifications are correctly translated into the spatial domain. To reconstruct the spatial domain image, \( S_{HH} \) is combined with the other sub-bands \( S_{LL}, S_{LH}, \) and \( S_{HL} \). For the SP embedded in \( S_{HH} \), the ISP is applied to reconstruct \( S_{HH} \) to spatial domain as follows:
\begin{equation}\label{eq:isp}
\footnotesize
    S_{HH}(x, y) = \sum_{u=0}^{L-1} \sum_{v=0}^{L-1} \zeta(u, v) \cdot \exp\left(i \frac{2 \pi}{L} \left(x \cdot u + y \cdot v\right)\right),
\end{equation}
where \( \zeta(u, v) \) represents spectral coefficients from the embedding process, \( L \) denotes the dimension of \( S_{HH} \), and \( (x, y) \) are spatial coordinates. SpecGuard then reconstructs the watermarked image \( I_{\text{embedded}} \) using the IWP as follows:
\begin{equation} \footnotesize \label{eq:iwp}
\begin{aligned}
    I_{\text{embedded}}(x, y) &= \text{IWP} (S_{LL}, S_{LH}, S_{HL}, S_{HH}).
\end{aligned}
\end{equation}

This process seamlessly embeds the watermark message \(M\) in the spectral domain, preserving the cover image \(I\)'s integrity. The inverse transformations that are expressed in~\cref{eq:isp} and~\cref{eq:iwp} fully restore visual quality, maintaining all frequency components.

\begin{algorithm}[!ht]
\footnotesize
\caption{\small SpecGuard decoder with wavelet, spectral projection with FFT approximation, and learnable threshold.}
\label{alg:decoding}
\begin{algorithmic}[1]

\STATE \textbf{Input:} Watermarked image \( I_{\text{embedded}} \), learnable \( \theta\), message length \( l \), radius \( r \), watermark channel \( W_{c} \)
\STATE \textbf{Output:} Decoded binary message \( D_{\text{M}} \)

\STATE \textbf{Procedure:} Apply Wavelet Projection on \( I_{\text{embedded}} \) to obtain \( S_{D_{LL}} \) (low-frequency) and \( S_{D_{HH}}^{\text{high}} \) (high-frequency)

\STATE \textbf{Procedure:} Spectral approximation with FFT (\( S_{D_{HH}}^{\text{high}} \)):
     \STATE \quad Separate even and odd indices: \( v = [x_{\text{even}}, \text{reverse}(x_{\text{odd}})] \)
    
     \STATE \quad Compute FFT on \( v \): \( V_{\text{complex}} = \text{FFT}(v) \)

    \STATE \quad \( V_{\text{real}} = V_{\text{complex}} \cdot \left[\cos\left(\frac{-\pi k}{2N}\right), \sin\left(\frac{-\pi k}{2N}\right)\right] \) \Comment  Calculate Real
    
    \STATE \quad \( V_{\text{real}}[0] \leftarrow \frac{V_{\text{real}}[0]}{\sqrt{N} \cdot 2}, V_{\text{real}}[1:] \leftarrow \frac{V_{\text{real}}[1:]}{\sqrt{\frac{N}{2}} \cdot 2} \) \Comment Energy preservation 
    
    \STATE \quad Transpose result and repeat to obtain \( S_{D_{HH}}^{\text{sp}} \) 
    
    \STATE \textbf{Return} \( S_{D_{HH}}^{\text{sp}} \)

\STATE \textbf{Procedure:} Pass \( S_{D_{HH}}^{\text{sp}} \) through sequential layers as:
    \[
    S_{D_{HH}}^{(n+1)} = \text{LeakyReLU} \left( \text{Conv}_{2D}\left(S_{HH}^{sp{(n)}}, K\right) \right), n = 1,...,k,
    \] 

\STATE \textbf{Return} \( S_{D_{HH}}^{(n+1)} \)

\STATE \textbf{Procedure:} Extraction (\( S_{D_{HH}}^{(n+1)}, l \)):
 \STATE \quad Set \( (\text{c}_x,\text{c}_y) = \left(\frac{H}{2}, \frac{W}{2}\right) \)
\STATE \quad Generate mask for high-frequency region within radius \( r \)
\item[] \quad \textbf{for} each coordinate \( (i, j) \) \textbf{do:}
    \STATE \quad \quad \( D(x_i, y_i) = \sqrt{(x_i - c_x)^2 + (y_i - c_y)^2} \) \Comment Euclidian Distance
    \STATE \quad \quad \textbf{if} \( D(x_i, y_i) \leq r \) \textbf{then}
        \STATE \quad \quad \quad Set \( \text{mask}[i,j] = 1 \)
    \STATE \quad \quad \textbf{end if}
\STATE \quad \textbf{end for}
\STATE \quad Extract mask: \( S_{D_{HH}}[:, W_{c}, \text{mask}[i,j]] \)
\STATE \quad Decode message using learnable \( \theta \):
    \[
    D_{\text{M}}[i] = 
    \begin{cases} 
    1 & \text{if } \text{Extracted}[i] > \theta\\  
    0 & \text{otherwise} 
    \end{cases} 
    \] 
\STATE \quad Update \( \theta \) dynamically: \( \theta \leftarrow \theta - \eta \cdot \frac{\partial L_{\text{dec}}}{\partial \theta} \) \Comment Optimizes robustness 

\STATE \textbf{Return} \( D_{\text{M}} \) 
\end{algorithmic}
\end{algorithm}

\subsection{Decoder}
\label{decoder}
As shown in~\cref{alg:decoding}, SpecGuard decoding process starts by applying wavelet projection (~\cref{eq:waveletprojection}) to the watermarked image \( I_{\text{embedded}} \), separating it into low and high-frequency bands, where the high-frequency band \( S_{D_{HH}}^{high} \) contains the embedded message similar to the process in the encoding phase, particularly in~\cref{eq:approximation}. An approximation of the spectral projection using FFT as shown in~\cref{eq:dct_cofficient} is then applied to \( S_{D_{HH}}^{\text{high}} \) returning the transformed data \( S_{D_{HH}}^{\text{sp}} \).
Then, \( S_{D_{HH}}\) is further refined through convolutional layers that captures the local features for message extraction.

To extract the message, a radial mask is created to isolate high-frequency areas within \( S_{D_{HH}}\), targeting the embedded regions based on their distance from the center. The masked values are compared against a learnable threshold \( \theta \) to decode each bit of the hidden message \( D_{\text{M}} \). Here, \( \theta \) serves as a threshold that adapts to the spectral patterns across the entire image, learning the distinct characteristics of the embedded watermark. From Parseval’s theorem~\cite{kelkar1983extension} ensures overall spectral and spatial energies remain equivalent, though local spectral energy distributions are altered by the watermark strength factor \(s\) .

The watermark’s strength factor \(s\) ensures that the high-energy areas where the message \(M\) is embedded as ``$1$'' remain robust, experiencing a minimum distortion in such conditions.
Moreover, this threshold can be optimized for better bit recovery accuracy during training. As \( \theta \) learns, it recognizes that areas encoded as ``$1$'' carry higher energy and impact due to the strength factor \(s\) of~\cref{strength}, while areas marked as ``$0$'', softened by the LeakyReLU’s minimal negative slope, have a lower intensity.  Such a dynamic approach enables \( \theta \) to identify and protect the embedded message \(M\) even when external disturbances occur, preserving the watermark’s structure within the watermarked image \( I_{\text{embedded}} \). And, \( \theta \) effectively learns to distinguish high-energy watermark regions. 
Therefore, the embedded message is more recoverable under diverse attacks, and SpecGuard's decoder ensures valid watermark bit extraction. Theoretical explanation of Parseval theorem's~\cite{kelkar1983extension} impact on message extraction is in the Supplementary. 
\subsection{Loss Calculation for SpecGuard}
To achieve the training objective of robust and invisible watermark embedding, a composite loss function is defined with two terms: encoder loss \( L_{\text{enc}} \) as expressed in~\cref{eq:encoder_loss} and decoder loss \( L_{\text{dec}} \) as expressed in~\cref{eq:decoder_loss}.
\begin{equation} \label{eq:encoder_loss}
\small
    \min_{\theta} \, \mathbb{E}_{(I, M) \sim D} \, L_{\text{enc}}(I,  I_\text{embedded}) = \| E_{\theta}(I, M) - I \|^2,
\end{equation}
\begin{equation} \label{eq:decoder_loss}
\small
    \min_{\theta} \, \mathbb{E}_{(I, M) \sim D} \, L_{\text{dec}}(M, D_{\text{M}}) = \| D_{\theta}(I_\text{embedded}) - M \|^2,
\end{equation}
where \( E_{\theta}(I, M) \) denotes the encoder output, embedding the message \( M \) into the cover image \( I \) to produce \( I_\text{embedded} \). By minimizing \( L_{\text{enc}} \), the encoder learns to embed the watermark invisibly, preserving the fidelity of the cover image. \( D_{\theta}(I_\text{embedded}) \) denotes the decoder’s output from the watermarked image \( I_\text{embedded} \). Minimizing \( L_{\text{dec}} \) enables the decoder to reliably retrieve the embedded message under varying conditions, such as noise and transformation.

The total loss \( L \) as shown in~\cref{eq:total_loss} used for optimizing the model combines these terms, balancing invisibility and robustness through weighted coefficients as follows:
\begin{equation} \label{eq:total_loss}
\small
\min_{\theta} L = \lambda_{\text{enc}} L_{\text{enc}} + \lambda_{\text{dec}} L_{\text{dec}},
\end{equation}
where \( \lambda_{\text{enc}} \) and \( \lambda_{\text{dec}} \) control the relative importance of visual fidelity and message recoverability. 
\section{Experimental Results}

\subsection{Dataset}
SpecGuard is trained on the MS-COCO dataset~\cite{lin2014microsoft}, which contains 25K images. To evaluate the robustness of the watermarking methods including our SpecGuard against different types of attacks, such as distortions, regenerations, and adversarial attacks, we used three datasets: DiffusionDB~\cite{wang2022diffusiondb}, MS-COCO~\cite{lin2014microsoft}, and DALL·E3~\footnote{https://huggingface.co/datasets/OpenDatasets/dalle-3-dataset}. Each of these datasets has a unique distribution of prompt words. We also ensured that no unethical or violent terms were included in the prompts. We randomly picked 200 images from  MS-COCO~\cite{lin2014microsoft} and applied watermark using SpecGuard for further verifying the robustness after uploading on various social media platforms and applying AI-based Photoshop Neural Filters (PNFs)~\footnote{https://www.adobe.com/products/photoshop/neural-filter.html}. The PNFs include depth blur, artistic style transfer, super zoom, JPEG artifact reduction, and colorization. For the super zoom filter, we set the `Sharpen' and `Noise Reduction' parameters to 15. For all other filters, we used the default settings.

\subsection{Implementation}
We used CUDA v11.3 and PyTorch with a batch size of 32 and the Adam optimizer on a multiple NVIDIA RTX 2080-equipped server. Mean Squared Error (MSE) and Bit Recovery Accuracy (BRA) are used for loss and accuracy calculation. We used Peak Signal-to-Noise Ratio (PSNR), Structural Similarity Index (SSIM), Fréchet Inception Distance (FID), and MSE to evaluate perceptual quality. Our model is trained for 300 epochs, with the decoder learning rate set to \(1 \times 10^{-3}\), reduced by half every 100 steps, and the encoder learning rate is set to \(1 \times 10^{-2}\) without scheduling. We set our watermark radius (\(r\)), strength factor (\(s\)), initial learning parameter (\(\theta\)), and the number of convolutional layers (\(k\)) to 100, 20, 0.001, and 32, respectively. This setup is applied with a message bit length (BL) of 48, 64, 128, and 256. Initially, decoder loss weight (\( \lambda_{\text{dec}} \)) and encoder loss weight (\( \lambda_{\text{enc}} \)) are set to 1.0 and 0.7, respectively. 
For assessing the robustness of watermarking methods against diverse attacks, we inherited the experimental setups from Waves~\cite{ding2024waves} and used effective metrics such as ``Quality at 95\% Performance (Q@0.95P)'', ``Quality at 70\% Performance (Q@0.7P)'', ``Avg P'' and ``Avg Q.'' Here, Q@0.95P and Q@0.7P indicate the level of image quality degradation required for watermark detection accuracy to reach 95\% and 70\%, respectively. The average performance (Avg P) metric represents the mean detection accuracy across various attack strengths, while the average quality degradation (Avg Q) measures the overall impact of attacks on image quality~\cite{fernandez2023stable,ding2024waves}. 

\begin{figure}[!t]
  \centering
   \includegraphics[width=0.9\linewidth]{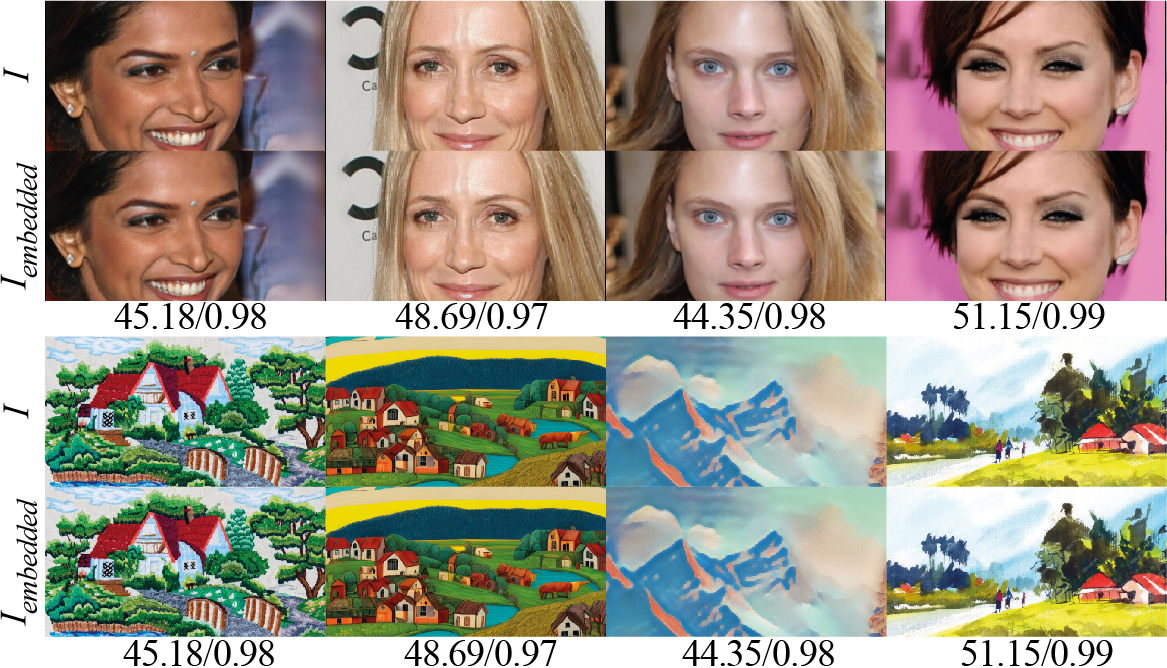}
   \caption{Some best results for cover vs watermarked images with PSNR/SSIM ($\uparrow$) scores showing minimal visual degradation when watermarked using proposed SpecGuard.}
   \label{fig:invisibility} 
\end{figure}

\begin{table} [!t]
    \centering
    \scriptsize
    \setlength{\tabcolsep}{0.9pt}
    \begin{tabular}{lcc|cc|cc}
        \hline 
         \multirow{2}{*}{\textbf{Metrics}} & \multicolumn{2}{c|}{\textbf{256 $\times$ 256}} & \multicolumn{2}{c|}{\textbf{512 $\times$ 512}} & \multicolumn{2}{c}{\textbf{1024 $\times$ 1024}} \\ \cline{2-3} \cline{4-5} \cline{6-7}
         & \textbf{CelebA-HQ} & \textbf{MS-COCO} & \textbf{CelebA-HQ} & \textbf{MS-COCO} & \textbf{CelebA-HQ} & \textbf{MS-COCO} \\ 
        \hline 
        PSNR$\uparrow$ & 40.361 & 40.320 & 44.651 & 44.680 & 48.170 & {48.081}  \\ 
        SSIM$\uparrow$ & 0.9889 & 0.9888 & 0.9927 & 0.9927 & 0.9937 & {0.9936}  \\
        FID$\downarrow$ & 16.451 & 16.690 & 16.972 & 17.020 & 17.446 & 16.955  \\
        MSE$\downarrow$ & 0.0002 & 0.0002 & 0.0001 & 0.0001 & 0.0001 & {0.0001}  \\ 
        \hline
    \end{tabular}
    \caption{Perceptual quality of SpecGuard watermarked images across resolutions and datasets using a fixed 30-bit message length.}
    \label{tab:invisibility}
\end{table}

\begin{table} [!t]
     \scriptsize
    \setlength{\tabcolsep}{2.5pt}
    \renewcommand{\arraystretch}{1.0}
    \begin{tabular}{l|llccccc}
    
        \hline 
           &\textbf{Methods} & \textbf{Venue} & \textbf{BL} & \textbf{PSNR}$\uparrow$ & \textbf{SSIM}$\uparrow$ & \textbf{FID}$\downarrow$ & \textbf{BRA}$\uparrow$ \\
        \hline

        \multirow{12}{*}{\textbf{\rotatebox{90}{Pre-processing methods}}} 
        & \multirow{3}{*}{Tree-Ring~\cite{NEURIPS2023_b54d1757}} & \multirow{3}{*}{NeurIPS'23} & 64 & 32.33 & 0.91 & 17.7 & 0.98 \\
        &&& 128 & 32.10 & 0.90 & 17.8 & 0.96 \\
        &&& 256 & 31.85 & 0.89 & 17.9 & 0.94 \\ \cline{2-8}

        & \multirow{3}{*}{Stable Signature~\cite{fernandez2023stable}} & \multirow{3}{*}{ICCV'23} & 64 & 30.00 & 0.89 & 19.6 & 0.98 \\
        &&& 128 & 29.80 & 0.88 & 19.7 & 0.96 \\
        &&& 256 & 29.50 & 0.87 & 19.8 & 0.96 \\ \cline{2-8}

        & \multirow{3}{*}{Yang et al.~\cite{yang2024gaussian}} & \multirow{3}{*}{CVPR'24} & 64 & 31.45 & 0.90 & 18.2 & 0.98 \\
        &&& 128 & 31.20 & 0.89 & 18.3 & 0.93 \\
        &&& 256 & 30.95 & 0.88 & 18.4 & 0.89 \\ \cline{2-8}

        & \multirow{3}{*}{SleeperMark~\cite{wang2024sleepermark}} & \multirow{3}{*}{CVPR'25} & 64 & 31.80 & 0.92 & 18.0 & 0.97 \\
        &&& 128 & 31.60 & 0.91 & 18.1 & 0.93 \\
        &&& 256 & 31.35 & 0.90 & 18.2 & 0.87 \\
        \hline \hline 

        \multirow{21}{*}{\textbf{\rotatebox{90}{Post-processing methods}}} 
        & \multirow{3}{*}{HiDDeN~\cite{hidden2018eccv}} & \multirow{3}{*}{ECCV'18} & 64 & 32.01 & 0.88 & 19.7 & 0.98 \\
        &&& 128 & 31.80 & 0.87 & 19.8 & 0.85 \\
        &&& 256 & 31.50 & 0.86 & 19.9 & 0.82 \\ \cline{2-8}

        & \multirow{3}{*}{StegaStamp~\cite{tancik2020stegastamp}} & \multirow{3}{*}{CVPR'20} & 64 & 28.50 & 0.91 & 17.9 & 0.99 \\
        &&& 128 & 28.20 & 0.90 & 18.0 & 0.98 \\
        &&& 256 & 28.00 & 0.89 & 18.1 & 0.94 \\ \cline{2-8}

        & \multirow{3}{*}{MBRS~\cite{jia2021mbrs}} & \multirow{3}{*}{ACM MM'21} & 64 & 38.20 & 0.96 & 17.9 & 0.98\\
        &&& 128 & 37.90 & 0.95 & 18.0 & 0.96 \\
        &&& 256 & 37.50 & 0.94 & 18.2 & 0.94 \\ \cline{2-8}

        & \multirow{3}{*}{FIN~\cite{fang2023flow}} & \multirow{3}{*}{AAAI'23} & 64 & 36.70 & 0.95 & 18.3 & 0.97\\
        &&& 128 & 36.40 & 0.94 & 18.4 & 0.96 \\
        &&& 256 & 36.10 & 0.93 & 18.5 & 0.96 \\ \cline{2-8}
        
        & \multirow{3}{*}{MuST~\cite{wang2024must}} & \multirow{3}{*}{AAAI'24} & 64 & 41.20 & 0.97 & 17.5 & 0.98\\
        &&& 128 & 40.90 & 0.96 & 17.6 & 0.93 \\
        &&& 256 & 40.50 & 0.95 & 17.8 & 0.90 \\ \cline{2-8}

        & \multirow{3}{*}{EditGuard~\cite{zhang2024editguard}} & \multirow{3}{*}{CVPR'24} & 64 & 41.56 & 0.97 & 17.8 & 0.98 \\
        &&& 128 & 41.30 & 0.96 & 17.9 & 0.97 \\
        &&& 256 & 40.90 & 0.95 & 18.0 & 0.97 \\ \cline{2-8}

        & \multirow{3}{*}{\textbf{SpecGuard (Ours)}} & \multirow{3}{*}{\textbf{ICCV'25}} & \cellcolor{customlightgray}64 & \cellcolor{customlightgray}\underline{42.59} & \cellcolor{customlightgray}\underline{0.98} & \cellcolor{customlightgray}\underline{17.2} & \cellcolor{customlightgray}0.99 \\ 
        &&& \cellcolor{customlightgray}128 & \cellcolor{customlightgray}\textbf{42.89} & \cellcolor{customlightgray}\textbf{0.99} & \cellcolor{customlightgray}\textbf{17.0} & \cellcolor{customlightgray}\textbf{0.99} \\
        &&& \cellcolor{customlightgray}256 & \cellcolor{customlightgray}40.86 & \cellcolor{customlightgray}\textbf{0.99} & \cellcolor{customlightgray}17.6 & \cellcolor{customlightgray}\underline{0.98} \\ \hline

        \multicolumn{8}{r}{\textit{\textbf{*BL}: Bit Length, \textbf{BRA:} Bit Recovery Accuracy}}\\
    \end{tabular}
    \caption{Comparison of SOTA pre-processing and post-processing watermarking methods with SpecGuard without attacks.}
    \label{tab:comparison}
    \vspace{-0.25cm}
\end{table}
\begin{table*}[!t]  
    \centering \scriptsize
    \renewcommand{\arraystretch}{1.1}
    \setlength{\tabcolsep}{0.1pt}
    \begin{tabular*}{\textwidth}{@{\extracolsep{\fill}}l lcccc|cccc|cccc|>{\columncolor{customlightgray}}c>{\columncolor{customlightgray}}c>{\columncolor{customlightgray}}c>{\columncolor{customlightgray}}c}  
        \hline 
        \multicolumn{2}{l}{\textbf{\multirow{2}{*}{Attack Type}}} & \multicolumn{4}{c|}{\textbf{Tree-Ring~\cite{NEURIPS2023_b54d1757}}} & \multicolumn{4}{c|}{\textbf{Stable Signature~\cite{fernandez2023stable}}} & \multicolumn{4}{c|}{\textbf{StegaStamp~\cite{tancik2020stegastamp}}} & \multicolumn{4}{c}{\cellcolor{customlightgray}\textbf{SpecGuard (Ours)}} \\ \cline{3-6} \cline{7-10} \cline{11-14} \cline{15-18} 
        
         & & \textbf{Q@0.95P} & \textbf{Q@0.7P} & \textbf{Avg P}& \textbf{Avg Q} & \textbf{Q@0.95P} & \textbf{Q@0.7P} & \textbf{Avg P} & \textbf{Avg Q}& \textbf{Q@0.95P} & \textbf{Q@0.7P} & \textbf{Avg P} & \textbf{Avg Q} & \textbf{Q@0.95P} & \textbf{Q@0.7P} & \textbf{Avg P} & \textbf{Avg Q} \\ \hline

        \multirow{11}{*}{\textbf{\rotatebox{90}{Distortions}}} & Rotation & 0.464 & 0.521 & 0.375 & 0.648 & 0.624 & 0.702 & 0.594 & 0.650 & 0.423 & 0.498 & 0.357 & 0.616 & 0.863 & 0.863 & 0.687 & 0.653\\ 
        
        & Crop & 0.592 & 0.592 & 0.332 & 0.463 & $\inf$ & $\inf$ & 0.995 & 0.461 & 0.602 & 0.602 & 0.540 & 0.451 & 0.812 & 0.812 & 0.998 & 0.742\\
  
        & Bright & $\inf$ & $\inf$ & $\inf$ & 0.304 & $\inf$ & $\inf$ & 0.998 & 0.305 & $\inf$ & $\inf$ & 0.998 & 0.317 & $\inf$ & $\inf$ & 0.998 & 0.466\\
        
        & Contrast & $\inf$ & $\inf$ & 0.998 & 0.243 & $\inf$ & $\inf$ & 0.998 & 0.243 & $\inf$ & $\inf$ & 0.998 & 0.231 & $\inf$ & $\inf$& 0.998 & 0.556\\
        
        & Blur & 0.861 & 1.112 & 0.563 & 1.221 & $-\inf$ & $-\inf$ & 0.000 & 1.204 & 0.848 & 0.962 & 0.414 & 1.000 & 0.921 & inf & 1.000 & 1.452\\
        
        & Noise & 0.548 & $\inf$ & 0.980 & 0.395 & 0.402 & 0.520 & 0.870 & 0.390 & $\inf$ & $\inf$ & 1.000 & 0.360 & inf & inf & 0.999 & 0.568\\
        
        & JPEG & 0.499 & 0.499 & 0.929 & 0.284 & 0.485 & 0.485 & 0.793 & 0.284 & $\inf$ & $\inf$ & 0.998 & 0.263 & inf & inf & 1.000 & 0.495\\
        
        & Geo & 0.525 & 0.593 & 0.277 & 0.768 & 0.850 & $\inf$ & 0.937 & 0.767 & 0.663 & 0.693 & 0.396 & 0.733 & 0.869 & 0.869 & 0.865 & 0.623\\
          
        & Deg & 0.620 & $\inf$ & 0.892 & 0.694 & 0.206 & 0.369 & 0.300 & 0.679 & 0.826 & 0.975 & 0.852 & 0.664 &  0.895 & 1.141 & 0.915 &  0.749\\
        
        & Combine & 0.539 & 0.751 & 0.403 & 0.908 & 0.538 & 0.691 & 0.334 & 0.900 & 0.945 & 1.101 & 0.795 & 0.870 & 0.979 & 1.256 & 0.911 & 0.952\\ \hline

    \multirow{6}{*}{\textbf{\rotatebox{90}{Regeneration}}}& Regen-Diff & $-\inf$ & 0.307 & 0.612 & 0.323 & $-\inf$ & $-\inf$ & 0.001 & 0.300 & 0.331 & $\inf$ & 0.943 & 0.327 &  inf & inf & 0.982 &  0.477\\
    
    & Regen-DiffP & $\inf$ & 0.307 & 0.601 & 0.327 & $-\inf$ & $-\inf$ & 0.001 & 0.303 & 0.333 & $\inf$ & 0.940 & 0.329 &  inf & inf &  0.982 & 0.562\\
    & Regen-VAE & 0.578 & 0.578 & 0.832 & 0.348 & 0.545 & 0.545 & 0.516 & 0.339 & $\inf$ & $\inf$ & 1.000 & 0.343 &  inf & inf & 0.995 &  0.521\\
    & Regen-KLVAE & $\inf$ & $\inf$ & 0.990 & 0.233 6 & $-\inf$ & 0.176 & 0.217 & 0.206 & $\inf$ & $\inf$ & 1.000 & 0.240 &  inf & inf & 0.990 &  0.492\\
    & Rinse-2xDiff & $-\inf$ & 0.333 & 0.510 & 0.357 & $-\inf$ & $-\inf$ & 0.001 & 0.332 & 0.391 & $\inf$ & 0.941 & 0.366 &  inf & inf & 0.993 &  0.561\\
    & Rinse-4xDiff & $-\inf$ & 0.355 & 0.443 & 0.466 & $-\inf$  & $-\inf$  & 0.000 & 0.438 & 0.388 & $\inf$ & 0.909 & 0.477 &  inf & inf & 0.992 &  0.533\\ \hline
    
     \multirow{8}{*}{\textbf{\rotatebox{90}{Adversarial}}}& AdvEmbG-KLVAE8 & $-\inf$ & 0.164 & 0.448 & 0.253 & $\inf$ & $\inf$ & 0.998 & 0.249 & $\inf$ & $\inf$  & 1.000 & 0.232 &  inf & inf & 1.000 &  0.456\\
    & AdvEmbB-RN18 & 0.241 & $\inf$ & 0.953 & 0.218 & $\inf$ & $\inf$ & 0.999 & 0.212 & $\inf$ & $\inf$ & 1.000 & 0.196 &  inf & inf & 1.000 &  0.467\\
    & AdvEmbB-CLIP & 0.541 & $\inf$ & 0.932 & 0.549 & $\inf$ & $\inf$ & 0.999 & 0.541 & $\inf$ & $\inf$ & 1.000 & 0.488 &  inf & inf & 1.000 &  0.436\\
    & AdvEmbB-KLVAE16 & 0.195 & $\inf$ & 0.888 & 0.238 & $\inf$ & $\inf$ & 0.997 & 0.233 & $\inf$ & $\inf$ & 1.000 & 0.206 &  inf & inf & 1.000 &  0.482\\
    & AdvEmbB-SdxlVAE &  0.222 & $\inf$ & 0.934 & 0.221 & $\inf$ & $\inf$ & 0.998 & 0.219 & $\inf$ & $\inf$ & 1.000 & 0.204 &  inf & inf & 1.000 &  0.492\\
    & AdvCls-UnWM\&WM & $-\inf$ & 0.102 & 0.499 & 0.145 & $\inf$ & $\inf$ & 0.999 & 0.101 & $\inf$ & $\inf$ & 1.000 & 0.101 &  inf & inf & 1.000 &  0.497\\
    & AdvCls-Real\&WM &  $\inf$ & $\inf$ & 1.000 & 0.047 & $\inf$ & $\inf$ & 0.998 & 0.092 & $\inf$ & $\inf$ & 1.000 & 0.106 &  inf & inf & 1.000 &  0.427\\
    & AdvCls-WM1\&WM2 & $-\inf$ & 0.101 & 0.492 & 0.139 & $\inf$ & $\inf$ & 0.999 & 0.084 & $\inf$ & $\inf$ & 1.000 & 0.129 &  inf & inf & 1.000 &  0.441\\\hline
     
    \end{tabular*}
    \caption{Robustness comparison various across attacks using Q@0.95P($\uparrow$), Q@0.7P($\uparrow$), Avg P($\uparrow$) and Avg Q($\uparrow$). Here, `inf' denotes that no attack was sufficient to degrade performance below the threshold, indicating strong robustness, whereas `-inf' signifies that even the weakest attack caused detection to fall below the threshold, reflecting weak robustness.}
    \label{tab:attacks}
\end{table*}

\begin{figure*}[!t]
  \centering
\includegraphics[width=0.95\linewidth]{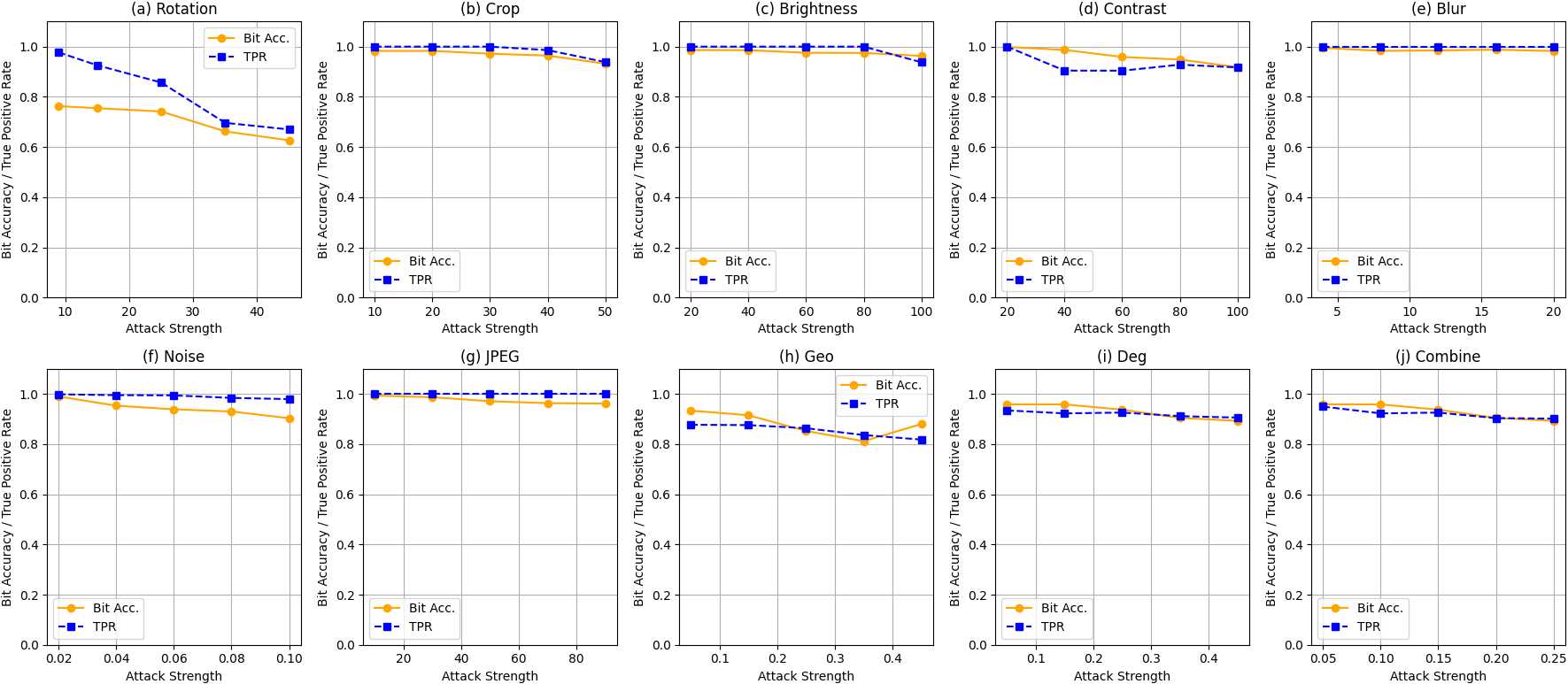}
    \caption{Robustness validation of our proposed SpecGuard under different distortion attacks, including geometric transformations: Geo (rotation, cropping), photometric modifications (brightness, contrast), and degradations: Deg (blur, noise, JPEG compression).}
   \label{fig:st3} 
   \vspace{-0.3cm}
\end{figure*}

\begin{table} [!ht]
     \scriptsize \centering
    \setlength{\tabcolsep}{1.2pt}
    \begin{tabular}{llcc|llcc}
        \hline 
        \multicolumn{2}{l}{\textbf{Modules}}  & \textbf{PSNR/SSIM$\uparrow$} & \textbf{BRA$\uparrow$} & \multicolumn{2}{l}{\textbf{Modules}}& \textbf{PSNR/SSIM$\uparrow$} & \textbf{BRA$\uparrow$}\\ \hline

        \multicolumn{2}{l}{WP($L_1$)} & 40.51/0.96 & 0.92 &  \multicolumn{2}{l}{\cellcolor{customlightgray} \textbf{WP($L_1$)+$ \text{SP}_{\text{$FA$}}$}} & \cellcolor{customlightgray}\textbf{42.89/0.99} & \cellcolor{customlightgray}\textbf{0.99} \\
        \multicolumn{2}{l}{WP($L_2$)} & 38.15/0.93 & 0.87 & \multicolumn{2}{l}{WP($L_2$)+$ \text{SP}_{\text{$FA$}} $} & 36.25/0.92 & 0.89 \\ 

        \hline \hline 
        \multicolumn{2}{l}{\textbf{Attacks}}  & \textbf{PSNR/SSIM$\uparrow$} & \textbf{BRA$\uparrow$} & \multicolumn{2}{l}{\textbf{Attacks}}& \textbf{PSNR/SSIM$\uparrow$} & \textbf{BRA$\uparrow$}\\ \hline

        \multicolumn{2}{l}{Rotate ($45^\circ$)} & 12.15/21.31 & 0.82 & \multicolumn{2}{l}{Rotate ($90^\circ$)} & 11.15/19.31 & 0.65 \\
        
        \multicolumn{2}{l}{Blur (0.3)} & 35.01/0.95 & 0.98 &   \multicolumn{2}{l}{Blur (0.6)} & 30.11/0.91 & 0.98 \\
        
        \multicolumn{2}{l}{Geo (0.3)} & 12.08/0.50 & 0.93 &   \multicolumn{2}{l}{Geo (0.6)} & 10.25/0.45 & 0.86 \\
     
        \hline
        \multicolumn{8}{r}{\textit{\textbf{*WP:} Wavelet Projection, \textbf{SP:} Spectral Projection, $FA$: FFT Approximation}}
    \end{tabular}
    \caption{Ablation studies on the proposed SpecGuard for across various configurations, setting \( M \) = 128, \(r\) = 100, and \(s\) = 20.}
    \label{tab:ablation}
    \vspace{-0.4cm}
\end{table}

\subsection{Watermark Invisibility}
\textcolor{black}{To evaluate the invisibility of the embedded watermark, we conducted perceptual and quantitative assessments using SpecGuard. As shown in~\cref{fig:invisibility}, there is no noticeable perceptual degradation between the cover and watermarked images, confirming that the watermark remains imperceptible to the human eye. For a more comprehensive evaluation, we created three subsets of different image sizes ranging between $256$ to $1024$ with images from the MS-COCO~\cite{lin2014microsoft} and CelebA-HQ~\cite{Karras2018CelebAHQ} datasets and applied the SpecGuard watermarking method to compare the average PSNR values between the cover and watermarked images, as in~\cref{tab:invisibility}.} 

\textcolor{black}{For quantitative evaluation, we further compare the performance of SpecGuard with the SOTA pre-processing and post-processing watermarking methods. As presented in~\cref{tab:comparison}, SpecGuard achieves the highest PSNR of 42.89 when the bit length was 128. Additionally, it attains the highest SSIM of 0.99 at a BL of 128 and 256 among all compared methods, indicating minimal visual distortion. Additionally, SpecGuard achieved the lowest FID of 17.0 and the highest BRA of 0.99, ensuring strong robustness while maintaining imperceptibility.
Overall, our results demonstrate that SpecGuard outperforms both pre-processing and post-processing watermarking methods, achieving superior imperceptibility and robustness.}

\subsection{Capacity}
To evaluate embedding capacity, we examined SpecGuard across different bit lengths and compared it with SOTA watermarking methods. Our experiments with 64, 128, and 256 bits demonstrate SpecGuard’s high capacity of bit embedding while maintaining perceptual quality and robustness, as shown in~\cref{tab:comparison}. Notably, it achieves a PSNR of 42.89, the highest among all methods, along with the highest BRA of 0.99 and the lowest FID of 17.0 at 128 bits, ensuring minimal visual impact. This adaptability to different bit lengths without quality loss makes SpecGuard ideal for applications requiring flexible watermark sizes. 
Unlike StegaStamp and HiDDeN, which suffer reduced BRA for higher message bits, SpecGuard consistently extracts bits across all tested lengths. SpecGuard’s theoretical watermark capacity is provided in the Supplementary.

\begin{table} [!t]
     \scriptsize \centering
    \setlength{\tabcolsep}{3.5pt}
    \begin{tabular}{lcc|lcc}
        \hline 
         \textbf{Platform} & \textbf{PSNR/SSIM$\uparrow$} & \textbf{BRA$\uparrow$} & \textbf{PS Filters} & \textbf{PSNR/SSIM$\uparrow$} & \textbf{BRA$\uparrow$}\\ 

        \hline 
        Facebook  & 48.56/0.97 & 0.97 & Depth Blur & 25.25/0.89 & 0.85\\
        LinkedIn  & 47.55/0.97 & 0.96 & StyleT. & 25.12/0.84 & {0.85}\\
        \underline{Instagram}  & \underline{48.56/0.98} & \underline{0.98} & \cellcolor{customlightgray} \textbf{Super Zoom}  & \cellcolor{customlightgray} \cellcolor{customlightgray} \textbf{36.15/0.88} & \cellcolor{customlightgray}\textbf{0.95} \\
        
        WhatsApp   & 42.10/0.96 & 0.97 & \underline{JPEG Artifacts}  & \underline{31.01/0.85} & \underline{0.94}\\
        
        \cellcolor{customlightgray}\textbf{X (Twitter)} & \cellcolor{customlightgray} \textbf{49.25/1.00} & \cellcolor{customlightgray} \textbf{0.99}  & Colorize & 23.15/0.82 & 0.92\\ 
        \hline
    \end{tabular}
    \caption{Evaluation of SpecGuard's robustness across Photoshop filters and while uploaded on different social media platforms.}
    \label{tab:social}
    \vspace{-0.4cm}
\end{table}

\subsection{Robustness}
\label{sec:robustness}
\textcolor{black}{We evaluate watermarking robustness by analyzing detection performance against a range of diverse and challenging real-world attacks.}
\textcolor{black}{Results demonstrate the strong robustness of SpecGuard across various attacks. For example, as presented in~\cref{tab:attacks}, against geometric distortions such as cropping and rotation, SpecGuard achieved an Avg P of 0.998 and 0.687, respectively. Similarly, across the combined distortion-based attacks, SpecGuard achieves an overall Avg P of $0.911$ and Avg Q of $0.952$, ensuring minimal quality loss while maintaining high detection accuracy. Notably, the high values of Q@0.95P and Q@0.7P indicate that SpecGuard can sustain reliable detection at strict performance thresholds, even under aggressive perturbations. Unlike prior methods that struggle with extreme transformations, SpecGuard shows remarkable robustness against regeneration-based attacks like Rinse-2xDiff~\cite{balle2018variational} (an image is noised then denoised by Stable Diffusion v1.4 two times with strength as a number of timesteps, 20-100) and Regen-VAE~\cite{balle2018variational}, maintaining high Avg P. Similarly, under adversarial attacks, SpecGuard consistently secures watermark detectability, outperforming existing techniques across all tested scenarios. These results establish SpecGuard as a highly robust watermarking approach capable of preserving image integrity even under severe distortions and adversarial manipulations, ensuring watermark reliability across diverse attack types. More details about how the attacks are performed are provided in supplementary material. Further, our results in~\cref{fig:st3} highlight the strong robustness of SpecGuard against various distortion attacks compared to other SOTA watermarking methods.}

\noindent
\textbf{Social Platforms and Photoshop Filters.}
SpecGuard's robustness when images are shared across social media platforms and subjected to common Photoshop Neural Filters (PNFs) is shown in~\cref{tab:social}. SpecGuard consistently maintains high PSNR and SSIM values, with BRA values close to 0.99 on platforms such as X (formally Twitter), Instagram, and Facebook. Also, it shown strong resilience to various PNFs, such as Super Zoom and JPEG Artifacts achieving BRA of 0.95 and 0.94. The PSNR, SSIM, and BRA values are expected to decrease with the severity of image manipulation, as increased manipulation leads to loss of image authenticity. For example, as we applied 60\% style transfer the PSNR and BRA decreased to 25.12 and 0.85. Similarly, the depth blur which excessively reduces the image clarity also causes the decrease of BRA to 0.85.

\subsection{Ablation Study}
We examined the impact of WP at different levels ($L_1$ and $L_2$) and its combination with SP using $FA$ in~\cref{tab:ablation}.
As observed, the WP($L_1$) + $ \text{SP}_{\text{$FA$}}$ configuration achieved the highest PSNR and SSIM values of 42.89 and 0.99, respectively, and BRA of 0.99, indicating improved watermark invisibility and robustness. In contrast, using WP alone at either $L_1$ or $L_2$ resulted in lower BRA, with values of 0.92 and 0.87, respectively, demonstrating that the combined WP + $ \text{SP}_{\text{$FA$}} $ approach significantly enhances performance. We also evaluated the robustness of SpecGuard under strong adversarial attacks identified in~\cref{tab:ablation}, such as rotation, blur, and geometric transformations. The results indicate that higher levels of attack severity, such as 90° rotation, lead to a more significant drop in PSNR, SSIM, and BRA, with values dropping to 11.15, 19.31, and 0.65, respectively. Despite this, the model shows relatively high resilience under moderate attack intensities, such as 45° rotation and low levels of blur and geometric distortion, achieving BRA values as high as 0.93 under geometric transformations at the 0.3 thresholds. 
More ablations are in the supplementary.

\section{Conclusion}
\label{sec:conclusion}

We propose SpecGuard, a novel invisible watermarking method that ensures secure and robust information concealment. Unlike traditional approaches, SpecGuard remains highly resilient against diverse distortions, adversarial attacks, and regeneration-based transformations. Experimental results demonstrate its superior bit recovery accuracy of 99\% maintaining high PSNR. By outperforming SOTA watermarking methods in both detection reliability and imperceptibility, SpecGuard establishes a new benchmark for watermarking under real-world constraints.

\section*{Acknowledgments}
This work was partly supported by Institute for Information \& communication Technology Planning \& evaluation (IITP) grants funded by the Korean government MSIT:
(RS-2022-II221199, RS-2022-II220688, RS-2019-II190421, RS-2023-00230337, RS-2024-00356293, RS-2024-00437849, RS-2021-II212068,  RS-2025-02304983, and  RS-2025-02263841).



{
    \small
    \bibliographystyle{ieeenat_fullname}
    \bibliography{main}
}
 \clearpage
\setcounter{page}{1}
\maketitlesupplementary
\begin{figure}[!t]
  \centering
   \includegraphics[width=1.0\linewidth]{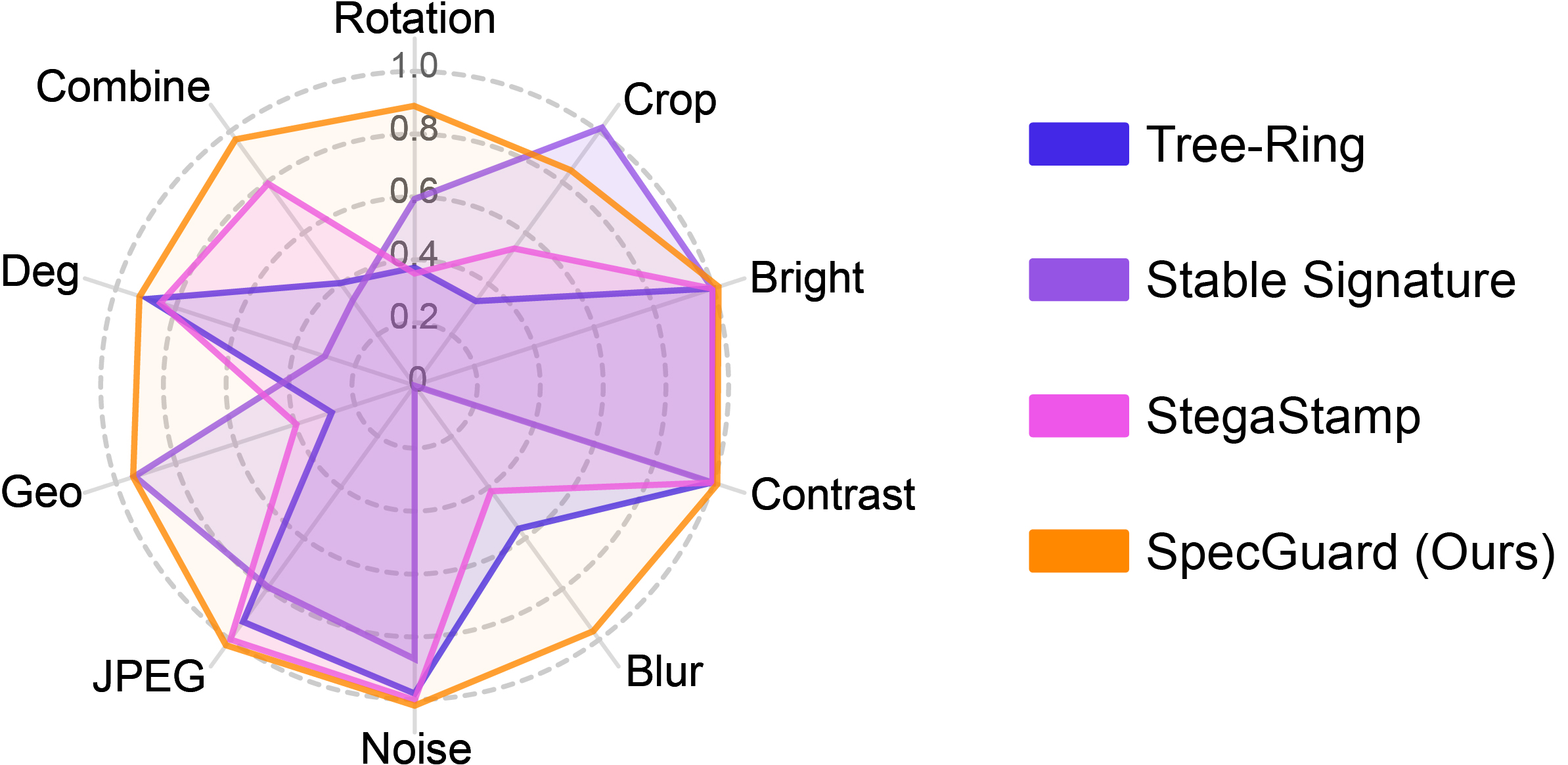}
   \caption{Comparison of SOTA watermarking methods in terms of average TPR@0.1\%FPR (90\% of watermarked images are correctly detected at 0.1\% false positive rate) under different attacks.}
   \label{fig:st4}
\end{figure}

\section{Summary of Notations}
\label{sec:notations}
To ensure clarity in understanding SpecGuard's mathematical formulation, we summarize the key notations used throughout the methodology (Sec.~{\textcolor{iccvblue}{3}}) of the main paper.
The complete set of notations is presented in~\cref{tab:notations}.

\section{Impact of Parseval's Theorem in Message Extraction}

To achieve robust and efficient decoding as detailed in Sec. \textcolor{iccvblue}{3.2} of the main paper, SpecGuard leverages Parseval's theorem~\cite{kelkar1983extension}, a fundamental principle in signal processing, which establishes energy equivalence between spatial and spectral domains. Formally, Parseval's theorem is defined as follows:
\begin{equation}
\sum_{x,y} |I(x,y)|^2 = \sum_{u,v} |\zeta(u,v)|^2,
\end{equation}
where \(I(x,y)\) denotes spatial-domain pixel intensities, and \(\zeta(u,v)\) represent their corresponding spectral-domain coefficients.

In SpecGuard, watermark embedding modifies selected spectral coefficients, introducing subtle local energy variations. The embedding process employs a strength factor \( s \), adjusting spectral energy differences as follows:
\begin{equation}
\zeta_{\text{embedded}}(u,v) = \zeta(u,v) + s \cdot W(u,v),
\end{equation}
where \(\zeta_{\text{embedded}}(u,v)\) denotes modified coefficients and \(W(u,v)\) is the spectral-domain watermark signal. Although local energy distribution is altered, the overall signal energy remains constant as guaranteed by Parseval's theorem as follows:
\begin{equation}
\sum_{x,y} |I_(x,y)|^2 = \sum_{u,v} |\zeta_{\text{embedded}}(u,v)|^2.
\end{equation}

During decoding, these local spectral energy variations, preserved due to total energy constancy, allow stable watermark extraction. Specifically, the decoder computes spectral projections via FFT approximation to isolate embedded spectral energy patterns as follows:
\begin{equation}
S_{D_{HH}}^{\text{sp}} = \text{SpectralProjectionFFT}(S_{D_{HH}}^{\text{high}}).
\end{equation}

The decoder subsequently employs a dynamically optimized threshold \( \theta \) to differentiate watermark signals from noise as follows:
\begin{equation}
D_{M}[i] = \begin{cases}
1 & \text{if } S_{D_{HH}}^{\text{sp}}[i] > \theta,\\[4pt]
0 & \text{otherwise}.
\end{cases}
\end{equation}

The adaptive threshold \( \theta \) is optimized via gradient descent during training, adapting to spectral energy distributions as follows:
\begin{equation}
\theta \leftarrow \theta - \eta \cdot \frac{\partial L_{\text{dec}}}{\partial \theta},
\end{equation}
where \( L_{\text{dec}} \) is the decoding loss, and \( \eta \) is the learning rate. Thus, Parseval’s theorem critically supports SpecGuard by preserving total spectral energy, enabling stable differentiation of watermark bits and reliable decoding even under diverse real-world image distortions and adversarial attacks.

\begin{table*}[ht]
    \centering  \small
    \setlength{\tabcolsep}{7.5pt}
    \renewcommand{\arraystretch}{1.2}
    \begin{tabular}{cl}
        \toprule
        \textbf{Notation} & \textbf{Description} \\
        \midrule
        \( I \) & Cover image \\
        \( I_{\text{embedded}} \) & Watermarked image \\
        \( M \) & Watermark message \\
        \( c \) & Number of channels (e.g., RGB has \( c=3 \)) \\
        \( H, W \) & Height and width of the image \\
        \( W(a,b) \) & Wavelet transform of signal \( f(x) \) \\
        \( a, b \) & Scaling and translation parameters in wavelet transform \\
        \( \psi \) & Mother wavelet function \\
        \(d)\) & Direction of each wavelet components derived from \( \psi \) \\
        \( \mathcal{\phi}(u, v), \mathcal{\psi}_H(u, v), \mathcal{\psi}_V(u, v), \mathcal{\psi}_D(u, v) \) & Every directional scaling and wavelet basis components \\
        \( S_{LL}, S_{LH}, S_{HL}, S_{HH} \) & Wavelet sub-bands (low and high frequency components) \\
        \( \beta_j \) & Feature set capturing frequency and spatial details \\
        \( \kappa \) & Decomposition level determined by image complexity \\
        \( T(x,y) \) & Pixel intensity in high-frequency sub-band \( S_{HH} \) \\
        \( \zeta(u,v) \) & Spectral projection coefficients \\
        \( s \) & Strength factor controlling embedding intensity \\
        \( (c_x, c_y) \) & Center coordinates of the image \\
        \( D(x_i, y_i) \) & Euclidean distance from the center \\
        \( r \) & Radius of embedding region \\
        \( W_c \) & Selected watermark channel for embedding \\
        \( \theta \) & Learnable threshold for watermark extraction \\
        \( F(u,v) \) & 2D Fast Fourier Transform (FFT) of the extended signal \\
        \( L_{\text{enc}}, L_{\text{dec}} \) & Encoder and decoder loss functions \\
        \bottomrule
    \end{tabular}
    \caption{Description of the notations we used in the Sec. \textcolor{iccvblue}{3} (main paper) to describe our proposed SpecGuard.}
    \label{tab:notations}
\end{table*}

\section{Mathematical Proof}
\subsection{Proof for $S_{HH}$ Band of Wavelet Projection.} Here we presented a proof of one of the wavelet projections $S_{HH}$ from Eq. (\textcolor{iccvblue}{4}) based on the Eq. (\textcolor{iccvblue}{6}) of the main paper.

\begin{flalign*}\scriptsize
\mathcal{\psi}_j^D(u) = 2^{j/2} \mathcal{\psi}^D(2^j u), 
\quad \text{//1D wavelet} &&
\end{flalign*}
\begin{flalign*}\scriptsize
\mathcal{\psi}_{j,m}^D(u) = 2^{j/2} \mathcal{\psi}^D(2^j u - m), 
\quad \text{//Translation}&&
\end{flalign*}
\begin{flalign*}\scriptsize
\mathcal{\psi}_{j,m,n}^D(u, v) = 2^{j/2} \mathcal{\psi}^D(2^j u - m) \cdot \mathcal{\psi}^D(2^j v - n), &&
\text{\scriptsize //2D wavelet}&&
\end{flalign*}
\begin{flalign*}\scriptsize
S_{HH}(j, m, n) = \int_{-\infty}^\infty \int_{-\infty}^\infty I(u, v) \cdot \\ \mathcal{\psi}_{j,m,n}^D(u, v) \, du \, dv, 
\quad \text{\scriptsize //Projection}&&
\end{flalign*}
\begin{flalign*}\scriptsize
S_{HH}(j, m, n) &= \int_{-\infty}^\infty \int_{-\infty}^\infty I(u, v) \cdot \left[ 2^{j/2} \mathcal{\psi}^D(2^j u - m) \right. \\
&\quad \left. \cdot \mathcal{\psi}^D(2^j v - n) \right] \, du \, dv, 
\quad \text{\scriptsize //Substitution} &&
\end{flalign*}
\begin{flalign*}\scriptsize
S_{HH}(j, m, n) &= \sum_{p=0}^{l-1} \sum_{q=0}^{l-1} T_{m,n} \cdot \mathcal{\psi}^D(2^j u - m) \\
&\quad \cdot \mathcal{\psi}^D(2^j v - n), 
\quad \text{\scriptsize //Discretization} &&
\end{flalign*}
\begin{flalign*}\scriptsize
W_\mathcal{\psi}^d(j, u, v) &= \frac{1}{l} \sum_{m=0}^{l-1} \sum_{n=0}^{l-1} T_{m,n} \cdot \mathcal{\psi}^D\left(m - u \cdot 2^{-j}, \right. \\
&\quad \left. n - v \cdot 2^{-j}\right), 
\quad \text{\scriptsize //Normalized} &&
\end{flalign*}

\begin{table} [!t]
     \footnotesize  \renewcommand{\arraystretch}{1.0}
    \setlength{\tabcolsep}{6.5pt}
    \begin{tabular}{lcccc}
        \toprule 
         \textbf{Activation Function} & \textbf{Radius (\( r \))} & \textbf{PSNR$\uparrow$} & \textbf{SSIM$\uparrow$} & \textbf{BRA$\uparrow$} \\ \midrule
         
          \multirow{3}{*}{ReLU}  & \( r \)(50) & 39.54 & 0.93 & 0.97 \\
            & \( r \)(75) & 38.64 & 0.91 & 0.93 \\
            & \( r \)(100) & 37.96 & 0.91 & 0.95 \\ \midrule

          \multirow{3}{*}{Tanh} & \( r \)(50) & 37.18 & 0.89 & 0.82 \\
            & \( r \)(75) & 35.33 & 0.85 & 0.78 \\
            & \( r \)(100) & 37.66 & 0.90 & 0.80 \\ \midrule

          \multirow{3}{*}{\textbf{LeakyReLU}} & \( r \)(50) & 39.77 & 0.96 & 0.98 \\
            & \( r \)(75) & 40.28 & 0.97 & 0.98 \\
           & \textbf{\( r \)(100)} & \textbf{42.89} & \textbf{0.99} & \textbf{0.99} \\ \bottomrule
          
    \end{tabular}
    \caption{Performance evaluation of SpecGuard for different radius size and activation functions  while the Strenth Factor is 20.}
    \label{tab:abl1}
\end{table}

\subsection{Maximum Theoretical Watermark Capacity}
\label{sec:watermark_capacity}

To determine the maximum theoretical watermark capacity of SpecGuard, we analyze the SpecGuard's embedding pipeline, which integrates wavelet projection and spectral projection. The capacity derivation considers three key stages: `wavelet projection,' `spectral projection,' and `watermark distribution,' with each stage affecting the number of available coefficients for embedding.

\noindent
\textbf{Impact of Wavelet Projection.} 
SpecGuard applies wavelet projection at decomposition level \( L \), dividing the image into sub-bands. The watermark is embedded in the high-frequency sub-band, which retains fine image details and ensures robustness against low-frequency distortions. The spatial dimensions of the wavelet sub-band are reduced by a factor of \( 2^L \) along both height and width, resulting in a down-sampling effect.

The number of available coefficients after wavelet decomposition is as follows:
\begin{equation}
N_{\text{WP}} = \frac{H \times W}{4^L},
\end{equation}
where \( H \) and \( W \) are the image height and width, respectively. Including all image channels \( c \), the total number of wavelet coefficients available for embedding is as follows:
\begin{equation}
N_{\text{WP,total}} = \frac{H \times W \times c}{4^L}.
\label{eq_sp}
\end{equation}

Thus, increasing the decomposition level \( L \) reduces the available spatial coefficients exponentially, limiting embedding capacity.

\noindent
\textbf{Impact of Spectral Project.}
SpecGuard employs spectral projection using FFT to distribute the watermark in the spectral domain. The spectral coefficients are selectively utilized based on an adaptive mask that prioritizes mid-to-high-frequency components while avoiding low frequencies (which contain most perceptual information) and extremely high frequencies (which are prone to compression loss).

The fraction of spectral coefficients selected for watermarking is denoted as \( f_{\text{spectral}} \) where spectral coefficients are used in between 20\% and 50\% as follows:
\begin{equation}
0.2 \leq f_{\text{spectral}} \leq 0.5.
\end{equation}

After spectral projection following~\cref{eq_sp}, the number of  coefficients available for embedding is as follows:

\begin{table} [!t]
     \footnotesize \renewcommand{\arraystretch}{1.0}
    \setlength{\tabcolsep}{4.0pt}
    \begin{tabular}{lcccc}
        \toprule 
         \textbf{Activation Function} & \textbf{Strength Factor (\( s \))} & \textbf{PSNR$\uparrow$} & \textbf{SSIM$\uparrow$} & \textbf{BRA$\uparrow$} \\
        \midrule 
        LeakyReLU  & \( s \)(5)  & 40.79 & 0.98 & 0.97 \\
        LeakyReLU  & \( s \)(10)  & 39.51 & 0.96 & 0.97  \\
        LeakyReLU  & \( s \)(15)  & 38.14 & 0.95 & 0.99 \\
        \textbf{LeakyReLU}  & \textbf{\( s \)(20)}  & \textbf{42.89} & \textbf{0.99} & \textbf{0.99} \\
        \bottomrule
    \end{tabular}
    \caption{Impact of Strength Factor for the best combination of the activation function (LeakyReLU) and radius \( r(100) \).}
    \label{tab:abl2}
\end{table}

\begin{figure*}[!ht]
  \centering
\includegraphics[width=1.0\linewidth]{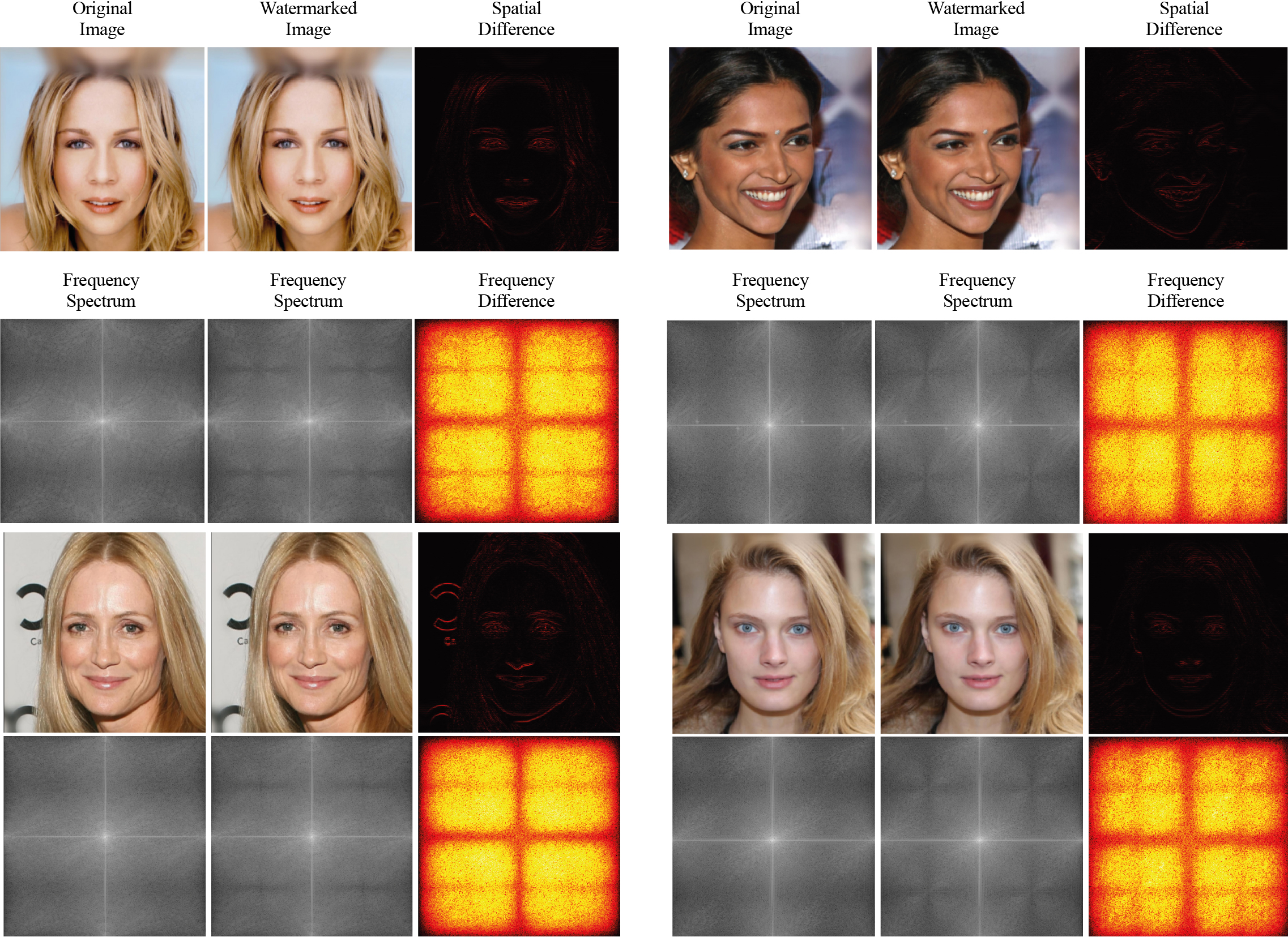}
   \caption{Visualization of the watermarking process using SpecGuard. The first row shows the original image, the watermarked image, and their spatial difference. The spatial difference highlights the minimal perceptual change between the original and watermarked images, ensuring imperceptibility. The second row presents the frequency spectrum of the original and watermarked images, along with their frequency difference, emphasizing the subtle embedding of the watermark in the high-frequency components. The comparison confirms that SpecGuard achieves invisible watermarking while maintaining robust frequency-domain characteristics for effective bit recovery.}
   \label{fig:st1}
\end{figure*}

\begin{figure}[!t]
  \centering
   \includegraphics[width=1.0\linewidth]{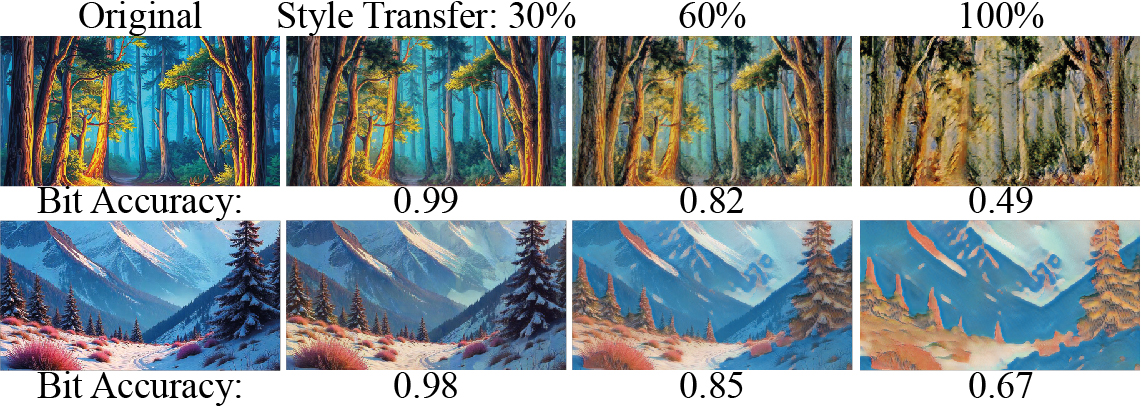}
   \caption{Effect of style transfer severity on bit recovery accuracy. As style intensity increases, bit accuracy decreases, showing the impact of major transformations.}
   \label{fig:styletransfer}
   \vspace{-0.4cm}
\end{figure}

\begin{figure}[!t]
  \centering
   \includegraphics[width=1.0\linewidth]{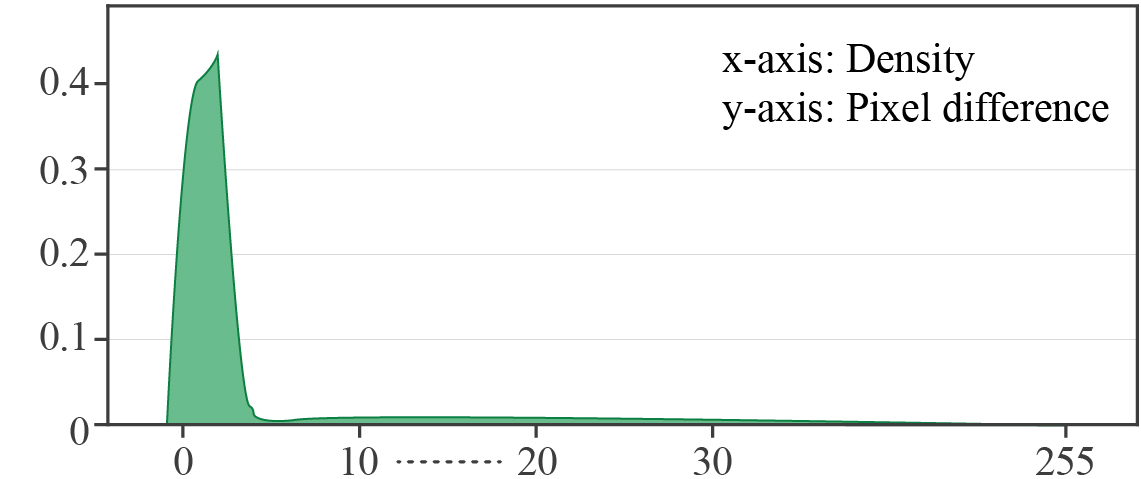}
   \caption{Pixel difference distribution between the original and watermarked images. The x-axis represents the pixel intensity difference, and the y-axis indicates the density. Most pixel differences remain close to zero, highlighting SpecGuard's minimal perceptual loss and superior imperceptibility of the embedded watermark.}
   \label{fig:distribution}
   \vspace{-0.4cm}
\end{figure}

\begin{equation} \small
N_{\text{SP}} = f_{\text{spectral}} \times N_{\text{WP,total}} = f_{\text{spectral}} \times \frac{H \times W \times c}{4^L}.
\end{equation}
A higher \( f_{\text{spectral}} \) increases embedding capacity but may reduce robustness to compression and noise, while a lower \( f_{\text{spectral}} \) focuses on the most resilient coefficients but limits capacity.

\noindent
\textbf{Watermark Distribution and Final Capacity.}
The watermark is distributed across the selected spectral coefficients \( f_{\text{spectral}} \) using a weighting scheme, where each coefficient can embed multiple bits. Let \( N_b \) represent the number of watermark bits per selected coefficient \( f_{\text{spectral}} \). The total embedded bits are then as follows:
\begin{equation}
C_{\text{total}} = N_b \times N_{\text{SP}}.
\end{equation}

Substituting \( N_{\text{SP}} \), the final maximum theoretical watermark capacity of SpecGuard is as follows:
\begin{equation} \small
C_{\max}(H, W, c, L, f_{\text{spectral}}, N_b) = \frac{H \times W \times c}{4^L} \times f_{\text{spectral}} \times N_b.
\end{equation}

The watermark capacity scales proportionally with the image dimensions \( H \times W \) and the number of channels \( c \), ensuring that larger images provide greater embedding space. However, higher wavelet decomposition levels \( L \) reduce the available capacity exponentially due to the \( 4^L \) down-sampling effect.
The fraction of spectral coefficients selected for embedding, denoted as \( f_{\text{spectral}} \), controls how much of the frequency domain is utilized, balancing capacity and robustness.
Additionally, the bit depth \( N_b \) determines the number of bits embedded per coefficient, directly influencing the total watermark payload.

Thus, SpecGuard achieves a flexible balance between capacity and robustness by leveraging adaptive spectral selection and wavelet decomposition, ensuring resilience under various transformations and attacks.

\begin{figure*}[!ht]
  \centering
\includegraphics[width=1.0\linewidth]{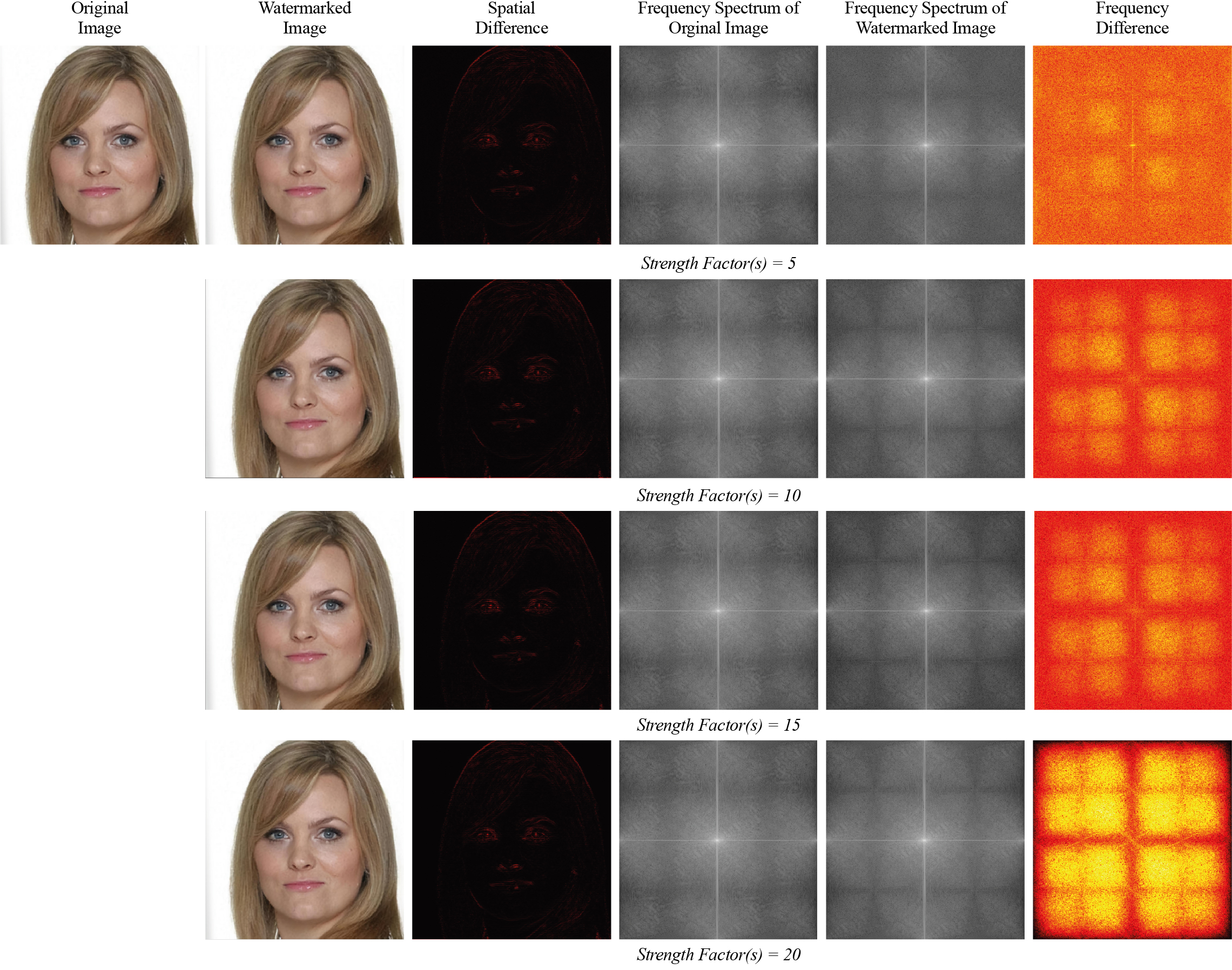}
   \caption{Visualization of the watermarking process using SpecGuard for different strength factors (\(s\)). The first row illustrates the original image, the watermarked image, and their spatial difference for \(s=5\), followed by the frequency spectra of the original and watermarked images and their frequency difference. The subsequent rows demonstrate the impact of increasing the strength factor (\(s=10, 15, 20\)) on the frequency difference, highlighting the progressive embedding intensity. Higher strength factors increase the visibility in the frequency domain while maintaining imperceptibility in the spatial domain, ensuring robust watermarking without compromising image quality.}
   \label{fig:st2}
\end{figure*}

\begin{table*}[!t]
    \centering
    \footnotesize
    \renewcommand{\arraystretch}{1.3}
    \setlength{\tabcolsep}{8pt}
    \begin{tabular}{p{2.5cm} p{7cm} p{6.0cm}}
        \hline
        \rowcolor{gray!25} \textbf{Attack Name} & \textbf{Description} & \textbf{Parameters} \\
        \hline\hline
        \multicolumn{3}{c}{\cellcolor{gray!10} \textbf{Distortion Attacks}} \\
        \hline
        Rotation & Rotates an image by a specified angle to test watermark robustness against geometric transformations. & Angle: 9° to 45° clockwise \\\hline
        Crop & Crops a portion of the image and resizes it back, simulating common editing. & Crop Ratio: 10\% to 50\% \\
        Bright & Adjusts image brightness to test watermark stability under illumination changes. & Brightness Increase: 20\% to 100\% \\\hline
        Contrast & Modifies image contrast to simulate lighting variations. & Contrast Increase: 20\% to 100\% \\\hline
        Blur & Applies a low-pass filter to smooth the image, reducing high-frequency details. & Kernel Size: 4 to 20 pixels \\\hline
        Noise & Introduces random pixel fluctuations to simulate compression noise and low-quality rendering. & Std. Deviation: 0.02 to 0.1 \\\hline
        JPEG & Compresses the image using JPEG encoding, reducing quality and adding artifacts. & Quality Score: 90 to 10 \\\hline
        
        Geo & Combination of geometric distortion attacks, & Strength: Geo(x): \\
        & including rotation, crop, applied uniformly & Rotation: \( 9^\circ + x \times (45^\circ - 9^\circ) \),\\
        
        &to assess cumulative effects.& Crop: \( 10\% + x \times (50\% - 10\%) \) \\\hline

        Deg & Combination of degradation attacks, integrating  & Strength: Deg(x):\\
        
        & blur, noise, and JPEG  & Blur: \( 4 + x \times (20 - 4) \), \\
        
        & to simulate complex real-world distortions. & Noise: \( 0.02 + x \times (0.1 - 0.02) \),\\
        
        & & JPEG: \( 90 - x \times (90 - 10) \) \\ \hline
        
        \hline
        \multicolumn{3}{c}{\cellcolor{gray!10} \textbf{Regeneration Attacks}} \\
        \hline
        Regen-Diff & Passes an image through a diffusion model to reconstruct a similar but altered version. & Denoising Steps: 40 to 200 \\\hline
        Regen-DiffP & A prompted version of diffusion-based regeneration, leveraging text guidance to refine results. & Denoising Steps: 40 to 200 with Prompt \\\hline
        Regen-VAE & Uses a variational autoencoder to encode and decode an image, affecting watermark integrity. & Quality Level: 1 to 7 \\\hline
        Regen-KLVAE & Uses a KL-regularized autoencoder to compress and reconstruct an image, weakening watermark signals. & Bottleneck Sizes: 4, 8, 16, 32 \\\hline
        Rinse-2xDiff & Applies a two-stage diffusion regeneration, progressively altering the image over multiple steps. & Timesteps: 20 to 100 per diffusion \\\hline
        Rinse-4xDiff & Performs four cycles of diffusion-based image reconstruction, aggressively erasing watermark traces. & Timesteps: 10 to 50 per diffusion \\\hline
        \hline
        \multicolumn{3}{c}{\cellcolor{gray!10} \textbf{Adversarial Attacks}} \\
        \hline
        AdvEmbG-KLVAE8 & Embeds adversarial perturbations using a grey-box VAE-based attack to reduce detection accuracy. & KL-VAE Encoding, $\epsilon$ = 2/255 to 8/255, PGD Iterations = 100, Step Size = 0.01$\times\epsilon$ \\\hline
        AdvEmbB-RN18 & Uses a pre-trained ResNet18 model to introduce adversarial noise and affect watermark recognition. & $\ell_{\infty}$ Perturbation: 2/255 to 8/255, PGD Iterations = 50, Step Size = 0.01$\times\epsilon$ \\\hline
        AdvEmbB-CLIP & Attacks the CLIP image encoder to introduce embedding shifts that disrupt watermark decoding. & $\ell_{2}$ Perturbation Norm = 2.5, PGD Iterations = 50, Learning Rate = 0.001 \\\hline
        AdvEmbB-KLVAE16 & Uses an alternative KL-VAE model to introduce structured perturbations into the embedding process. & KL-VAE Embedding, Latent Size = 16, $\ell_{\infty}$ Perturbation = 4/255 \\\hline
        AdvEmbB-SdxlVAE & Attacks Stable Diffusion XL’s VAE encoder to alter latent representations and remove watermarks. & Targeted VAE Perturbation, Diffusion Steps = 100, $\ell_{2}$ Perturbation = 3.0 \\\hline
        AdvCls-UnWM\&WM & Trains a surrogate detector on watermarked and non-watermarked images to bypass watermark detection. & Dataset Size = 3000 Images (1500 Per Class), ResNet-18, Learning Rate = 0.001, Batch Size = 128 \\\hline
        AdvCls-Real\&WM & Trains an adversarial classifier using real and watermarked images to classify watermark presence. & Dataset Size = 15,000 Images (7500 Per Class), Adam Optimizer, Learning Rate = 0.0005, Batch Size = 128, Epochs = 10 \\\hline
        AdvCls-WM1\&WM2 & Exploits watermark signal variations between different users to remove or alter hidden information. & Two Sets of Watermarked Images, Model = Vision Transformer (ViT), PGD Attack, Perturbation Strength = 6/255 \\
        \hline
    \end{tabular}
    \caption{Overview of attack types, their mechanisms, and key parameters based on the prior study~\cite{ding2024waves} that we also utilized in our study.}
    \label{tab:attack_summary}
\end{table*}

\section{Impact of Hyperparamters}
The performance of SpecGuard is influenced by several key hyperparameters, including the activation function, radius size (\( r \)), and strength factor (\( s \)). Each parameter plays a vital role in balancing the trade-off between perceptual quality, robustness, and watermark recovery accuracy. In addition to the ablation studies shown in Section 4.5 in the main paper, here we analyze the effect of the hyperparameters individually by conducting experiments under controlled conditions and report the findings in~\cref{tab:abl1} and~\cref{tab:abl2}. All the experiments presented here were conducted using a 128-bit watermark message.

\subsection{Activation Function and Radius} 
Table~\ref{tab:abl1} highlights the performance of SpecGuard with various activation functions, including ReLU~\cite{glorot2011deep}, Tanh~\cite{rumelhart1986learning}, and LeakyReLU~\cite{xu2015empirical}, while keeping the strength factor \( s \) fixed at 20. Among these, LeakyReLU outperforms others in terms of PSNR, SSIM, and bit recovery accuracy values across different radius sizes. Notably, with a radius \( r \) of $100$, LeakyReLU achieves a PSNR and SSIM of 42.89 and 0.99, respectively, with a bit recovery accuracy of 0.99. Overall, the results indicate the effectiveness of LeakyReLU for robust and invisible watermarking compared to ReLU and Tanh. While testing with different \( r \), such as 50 and 75, we observed a slightly lower perceptual quality and bit recovery accuracy. Therefore, we propose the SpecGuard with a combination of LeakyReLU, \( r \) of $100$ and \( s \) of 20.


\subsection{Strength Factor}

Table~\ref{tab:abl2} investigates the impact of the strength factor (\( s \)) using the best combination of LeakyReLU and radius \( r(100) \). A strength factor of \( s(20) \) achieves optimal performance with a PSNR/SSIM of 42.89/0.99 and a BRA of 0.99. Increasing \( s \) beyond 20 reduces PSNR and SSIM values, indicating diminished perceptual quality, while lower strength factors compromise robustness. Therefore, \( s(20) \) effectively balances robustness and visual quality as also shown in~\cref{fig:st1}.

Figure~\ref{fig:st2} further demonstrates the effect of different strength factors (\( s = 5, 10, 15, 20 \)) on the watermark embedding process. The first row showcases the original image, the watermarked image, and their spatial difference, highlighting the imperceptibility of the watermark in the spatial domain. The subsequent rows compare the frequency spectrum of the original and watermarked images, as well as the frequency difference, illustrating how increased strength factors enhance the visibility of the watermark in the frequency domain while maintaining imperceptibility in the spatial domain. Illustrate the robustness and adaptability of the proposed SpecGuard model in embedding and retaining watermark information under varying conditions.

\begin{table*}[!t]
    \footnotesize
    \renewcommand{\arraystretch}{1.1}
    \setlength{\tabcolsep}{9pt}
    \centering
    \caption{Robustness comparison of SpecGuard component configurations under four common perturbations: horizontal/vertical flip, downscaling (0.75×), and saturation increase (+40\%). We report PSNR and Bit Recovery Accuracy (BRA) under each condition. The full configuration (WP + SP + adaptive $\theta$) consistently achieves the highest robustness and fidelity across all settings, demonstrating the complementary benefits of spectral-domain embedding and adaptive thresholding.}
    \begin{tabular}{lcc|cc|cc|cc}
        \toprule
        \multirow{2}{*}{\textbf{Config}} & \multicolumn{2}{c|}{\textbf{No Attack}} & \multicolumn{2}{c|}{\textbf{Flip (avg. H/V)}} & \multicolumn{2}{c|}{\textbf{Scale 0.75×}} & \multicolumn{2}{c}{\textbf{Satur +40\%}} \\
        \cmidrule(lr){2-3} \cmidrule(lr){4-5} \cmidrule(lr){6-7} \cmidrule(lr){8-9}
        & \textbf{PSNR$\uparrow$} & \textbf{BRA$\uparrow$} 
        & \textbf{PSNR$\uparrow$} & \textbf{BRA$\uparrow$} 
        & \textbf{PSNR$\uparrow$} & \textbf{BRA$\uparrow$} 
        & \textbf{PSNR$\uparrow$} & \textbf{BRA$\uparrow$} \\
        \midrule
        WP (fixed $\theta$)              & 35.3$\pm$0.4 & 0.92$\pm$0.01 &  9.2$\pm$0.5 & 0.25$\pm$0.05 & 31.2$\pm$0.3 & 0.65$\pm$0.03 & 29.3$\pm$0.4 & 0.63$\pm$0.03 \\
        SP (fixed $\theta$)              & 36.6$\pm$0.4 & 0.93$\pm$0.01 & 11.5$\pm$0.6 & 0.33$\pm$0.05 & 32.6$\pm$0.3 & 0.70$\pm$0.03 & 30.8$\pm$0.4 & 0.68$\pm$0.03 \\
        WP + SP (fixed $\theta$)         & 38.8$\pm$0.3 & 0.93$\pm$0.01 & 13.9$\pm$0.5 & 0.48$\pm$0.04 & 34.2$\pm$0.3 & 0.85$\pm$0.02 & 32.8$\pm$0.3 & 0.83$\pm$0.02 \\
        \midrule
        \textbf{WP + SP + $\theta$ (Full)} & \textbf{42.9$\pm$0.2} & \textbf{0.99$\pm$0.005} & \textbf{16.2$\pm$0.4} & \textbf{0.60$\pm$0.04} & \textbf{35.3$\pm$0.3} & \textbf{0.94$\pm$0.02} & \textbf{34.6$\pm$0.3} & \textbf{0.92$\pm$0.02} \\
        \bottomrule
        \multicolumn{9}{r}{\textit{Mean $\pm$ standard deviation\ over three random seeds per configuration. \textbf{WP:} Wavelet Projection, \textbf{SP:} Spectral Projection.}} \\
    \end{tabular}
    \label{tab:robustness_re}
\end{table*}

\begin{figure}[t]
  \centering
  \begin{subfigure}[t]{0.48\linewidth}
    \includegraphics[width=\linewidth]{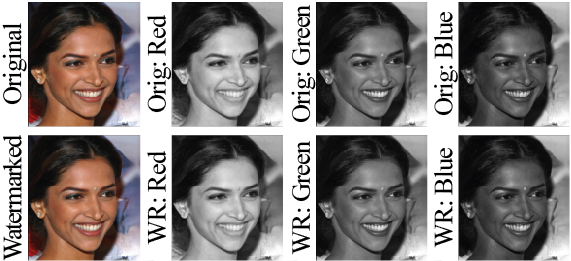}%
  \end{subfigure}
  \hfill
  \begin{subfigure}[t]{0.48\linewidth}
    \includegraphics[width=\linewidth]{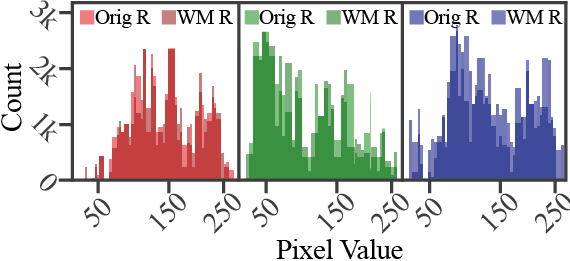}%
  \end{subfigure}
   \caption{$\Delta R/G/B$ maps of original vs.\ watermarked images.}
  \label{fig:rgb_maps}
\end{figure}

\section{Visual Analysis of Watermarked Images}

To further support the claim of imperceptibility, we provide a visual and channel-wise analysis of the original and watermarked images in Fig.~\ref{fig:rgb_maps}. The left panel shows side-by-side comparisons of the original and watermarked versions, along with their individual R, G, and B channels. The differences are visually negligible, indicating minimal perceptual impact from the embedding process.

The right panel presents histograms of pixel intensities for each color channel before and after watermarking. The distributions of red, green, and blue intensities remain highly consistent between the original and watermarked images. These results validate that SpecGuard preserves low-level color statistics and visual fidelity across all channels, aligning with the high PSNR and SSIM values reported in the main paper.

\section{Additional Ablation Studies}

To further understand the contribution of each component of SpecGuard, we conducted ablation experiments across different architectural configurations and evaluated their robustness under a range of perturbations, including horizontal/vertical flips, downscaling, and saturation increase. The results are summarized in Tab.~\ref{tab:robustness_re}.

Applying only Wavelet Projection (WP) or Spectral Projection (SP) with a fixed threshold provides moderate robustness under distortions such as flip (BRA = 0.25–0.33) and scaling (BRA = 0.65–0.70). Combining WP and SP without a learnable threshold further improves recovery, particularly under geometric distortions (e.g., Flip BRA = 0.48, Scale BRA = 0.85).

The full configuration of SpecGuard, which includes WP, SP, and a learnable threshold $\theta$ guided by Parseval's theorem, achieved the highest robustness across all categories. For instance, under flip perturbations, the BRA improved from 0.48 to 0.60. Similarly, under saturation enhancement, the BRA improved from 0.83 to 0.92. Notably, this improvement was achieved while maintaining high fidelity under no attack (PSNR = 42.9 ± 0.2, BRA = 0.99 ± 0.005).

These results confirm the complementary roles of wavelet-domain localization and spectral-domain embedding, with the adaptive threshold $\theta$ enabling reliable bit recovery under challenging distortions. Overall, the full SpecGuard architecture balances imperceptibility and robustness more effectively than any other partial configuration.

\section{Description of Benchmarking Attacks}

To comprehensively evaluate watermark robustness, we benchmark performance against a diverse set of attacks, including distortions, regeneration, and adversarial manipulations. These attacks, derived from prior benchmarking efforts~\cite{ding2024waves}, assess the stability of watermarks under real-world transformations. The results are presented in Tab.~\textcolor{iccvblue}{3} (main paper) and the details of the attacks are in~\cref{tab:attack_summary}, comparing multiple state-of-the-art (SOTA) methods such as Tree-Ring~\cite{NEURIPS2023_b54d1757}, Stable Signature~\cite{fernandez2023stable}, and StegaStamp~\cite{tancik2020stegastamp}. The attacks are categorized as follows:

\subsection{Distortion Attacks} These include standard image-processing transformations that alter the spatial or color properties of images. We consider rotation (9° to 45°) where images are rotated at varying degrees to test watermark stability. Resized cropping (10\% to 50\%) removes portions of an image and resizes the remaining content, mimicking common real-world editing. Random erasing (5\% to 25\%) replaces regions with gray pixels, simulating object removal. Brightness adjustments (20\% to 100\%) and contrast modifications (20\% to 100\%) simulate lighting variations. Gaussian blur (4 to 20 pixels) applies low-pass filtering, while Gaussian noise (0.02 to 0.1 standard deviation) adds random pixel fluctuations, simulating compression noise~\cite{ding2024waves}.

\subsection{Regeneration Attacks} These attacks leverage generative models such as diffusion and variational autoencoders (VAEs) to reconstruct images while suppressing embedded watermarks. We evaluate single regeneration attacks including Regen-Diff (diffusion-based reconstruction), Regen-DiffP (perceptually optimized diffusion), Regen-VAE (autoencoder-based reconstruction), and Regen-KLVAE (KL-regularized VAE reconstruction). Additionally, multi-step regeneration attacks such as Rinse-2xDiff and Rinse-4xDiff involve iterative diffusion processes designed to further erase watermark traces~\cite{saberi2023robustness, zhao2025invisible}.

\subsection{Adversarial Attacks} These attacks attempt to deceive watermark detectors through embedding perturbations or surrogate model training. Grey-box embedding attacks (AdvEmbG-KLVAE8) perturb watermarks while preserving image content. Black-box embedding attacks (AdvEmbB-RN18, AdvEmbB-CLIP, AdvEmbB-KLVAE16, AdvEmbB-SdxlVAE) introduce noise during watermark embedding to decrease detection confidence. Adversarial classifiers (AdvCls-UnWM\&WM, AdvCls-Real\&WM, AdvCls-WM1\&WM2) use learned classifiers to distinguish watermarked images and remove hidden signals~\cite{he2016deep,radford2021learning, podell2023sdxl, saberi2023robustness}.

Overall, our evaluation framework ensures a rigorous assessment of watermark robustness under various real-world transformations and adversarial strategies.

\section*{Acknowledgments}
This work was partly supported by Institute for Information \& communication Technology Planning \& evaluation (IITP) grants funded by the Korean government MSIT:
(RS-2022-II221199, RS-2022-II220688, RS-2019-II190421, RS-2023-00230337, RS-2024-00356293, RS-2024-00437849, RS-2021-II212068,  RS-2025-02304983, and  RS-2025-02263841).


\end{document}